# Towards Federated Multi-Armed Bandit Learning for Content Dissemination using Swarm of UAVs

AMIT KUMAR BHUYAN

Michigan State University, bhuyanam@msu.edu

HRISHIKESH DUTTA

Michigan State University, duttahr1@msu.edu

SUBIR BISWAS

Michigan State University, sbiswas@msu.edu

This paper introduces an Unmanned Aerial Vehicle - enabled content management architecture that is suitable for critical content access in communities of users that are communication-isolated during diverse types of disaster scenarios. The proposed architecture leverages a hybrid network of stationary anchor UAVs and mobile Micro-UAVs for ubiquitous content dissemination. The anchor UAVs are equipped with both vertical and lateral communication links, and they serve local users, while the mobile micro-ferrying UAVs extend coverage across communities with increased mobility. The focus is on developing a content dissemination system that dynamically learns optimal caching policies to maximize content availability. The core innovation is an adaptive content dissemination framework based on distributed Federated Multi-Armed Bandit learning. The goal is to optimize UAV content caching decisions based on geo-temporal content popularity and user demand variations. A Selective Caching Algorithm is also introduced to reduce redundant content replication by incorporating inter-UAV information sharing. This method strategically preserves the uniqueness in user preferences while amalgamating the intelligence across a distributed learning system. This approach improves the learning algorithm's ability to adapt to diverse user preferences. Functional verification and performance evaluation confirm the proposed architecture's utility across different network sizes, UAV swarms, and content popularity patterns.

## 1 INTRODUCTION

Earthquakes, floods, wars, and other catastrophic events can wreak havoc on human lives due to destruction of property and essential infrastructure such as those used for cellular communication. Destruction of communication infrastructure can leave an entire community cut off from vital information sources which can add to other disaster related difficulties. This paper proposes to employ Micro-Unmanned Aerial Vehicles (Micro-UAVs) as an innovative content delivery solution when traditional communication infrastructure including cell towers becomes inaccessible.

The rise of miniaturized UAV technology has made Micro-UAVs increasingly viable and cost-effective due to their small form factor and reduced power consumption compared to larger UAVs [1]. Their ability to fly at low altitudes makes them well-suited for unimpeded short-range communication. These advantages motivate the development of a UAV-assisted content dissemination system that can deploy a large fleet of cost-effective Micro-UAVs [2] when fixed communication infrastructure is not available. While being agile, the Micro-UAVs do come with their own challenges, including limited storage, battery, and flight capabilities. These introduce specific complexities into the content storage and dissemination process, which this paper attempts to address. This paper introduces a Federated Multi-armed Bandit (FedMAB) Learning based approach in order to optimize content caching in communication-constrained environments. The approach leverages explicit model sharing across UAVs, and employs a multi-dimensional reward structure to learn user content request patterns as experienced by the individual UAVs. Each UAV aims to maximize cumulative rewards [3] to improve its respective caching decision, thereby improving overall caching decisions and content dissemination via model sharing [4].

The framework spans scenarios in which disaster-affected populations are geographically clustered into isolated communities without access to surviving cell towers. In such circumstances, content request patterns and tolerable access delays ($TAD$) [5] may vary based on their urgency. The proposed solution deploys adaptive learning and on-the-fly caching decisions by the UAVs to deal with user requests with variable $TAD$s, and that is without prior knowledge of content request patterns. The content provisioning system adopts a two-tier architecture, comprising of relatively larger anchor-UAVs (A-UAVs), and small Micro-ferrying-UAVs (MF-UAVs). Each disaster-stuck isolated community is served by an A-UAV with its high-cost satellite-based vertical links such as the ones provided by Starlink ISP [6]. The MF-UAVs do not possess such vertical links. Their sole purpose is to algorithmically distribute content across those A-UAVs in manner that improves content availability to the isolated communities of users. It is worth noting that special purpose ground vehicles with similar communication equipment can also serve the role of A-UAVs. The overall objective is to ensure high content availability access to all communities, and that is while minimizing the utilization of the expensive vertical links on the A-UAVs. To achieve this, the paper addresses two key questions: First, what content caching policies should be implemented for both A-UAVs and the MF-UAV swarms to maximize content availability? Second, which content should be transferred from A-UAVs to MF-UAVs and when in order to support those caching policies? The proposed policy addresses these questions through on-the-fly policy learning with a FedMAB learning arrangement.

Existing research [7]-[27] on UAV-based content provisioning has emphasized on high-performance communication equipment that adds weight to UAVs, thus leading to rapid power consumption and limiting their operation duration and other capabilities. Furthermore, the cost of larger UAVs restricts their usage. The proposed architecture addresses these limitations by introducing a hierarchy in a which a swarm of low-cost Micro-UAVs are used as content carriers between a set of fewer relatively larger anchor A-UAVs. This eliminates the need for high-performance long-distance communication, leading to low operational energy budgets. In addition, due to a relatively large number of MF-UAVs in a swarm, the architecture can tolerate failure of up to a few of them while maintaining the system performance within an acceptable range. This has the potential for desirable failure tolerance in the proposed architecture.

The key contributions of the paper are as follows. First, a content dissemination system using swarm of Micro-UAVs is designed for on-demand content dissemination in a communication challenged environment. Second, a *Top-k* Multi-armed Bandit based method is deployed for on-the-fly learning of optimal caching policies in UAVs. A multi-dimensional reward structure for the *Top-k* MAB model is developed based on shared information via micro-UAVs. These rewards take local and global context of content popularities into consideration while learning optimal caching policies. Third, a novel model sharing approach is proposed via FedMAB Learning-based caching policy that considers the similarities among request patterns to achieve federation. This method is robust to lag in information while sharing models for FedMAB Bandit Learning. Fourth, a *selective caching algorithm* is designed that jointly, with FedMAB Learning-based Caching strategy, decides the caching policies of Micro-UAVs to manage the trade-off between effective caching capacity and UAV accessibility. Fifth, the interactions between learnt caching policies and Quality-of-Service (QoS) expectation, namely,

Tolerable Access Delay, is studied and characterized. Finally, simulation experiments and analytical models are developed for functional verification and performance evaluation of the proposed framework.

## 2 RELATED WORK

Extensive research has been conducted on micro-UAV applications, like in imaging, surveillance, terrestrial imaging, and precision agriculture [1]. In precision agriculture, autonomous micro-UAVs have been explored for yield estimation, crop fertilization, and monitoring [2], that demonstrates capabilities in image acquisition and processing for autonomous landing and maneuvering [7]. The Det-Fly database introduced in [8] has been crucial for training deep neural networks, enabling vision-based micro-UAV swarming, malicious UAV detection, and collision avoidance [8]. This paper investigates the potential of micro-UAVs for content dissemination, adopting an algorithm-centric approach to cache optimization.

In recent years, substantial research has broadly categorized UAV-caching into platform enhancements and algorithmic optimizations. In terms of platform enhancements, studies in [5] show that the effective caching capacity of UAVs can be improved by using solid-state drives (SSDs) due to their higher storage density and lower power consumption. The study in [9] demonstrates how caching capacity is enhanced by increasing the communication range between UAVs and ground nodes. Another study [10] suggests that UAVs flying at higher altitudes can cover larger areas, thus increasing the effective caching capacity of the system. It also discusses a multi-UAV caching strategy utilizing UAVs at different altitudes to optimize caching capacity and coverage for specific applications. Authors in [11, 12] implement energy-aware multi-armed bandit algorithms to identify user hotspots, maximizing data transmission rates without incurring excessive UAV flight and hover energy expenditures. While these approaches primarily focus on platform-related enhancements, they differ from the objectives of this article, which addresses cache optimization with an algorithm-centric perspective.

Another crucial aspect of UAV-assisted caching is energy efficiency, which directly influences system longevity and operational feasibility. While our approach indirectly manages UAV energy by optimizing caching strategies to reduce unnecessary mobility, complementary research has explored direct energy replenishment techniques. The work in [42] investigates the use of Unmanned Ground Vehicles (UGVs) as mobile charging stations which employs reinforcement learning to optimize UAV recharging schedules and flight trajectories. While this study focuses on energy management, our work primarily addresses the combinatorial complexity of content selection for caching, a challenge that arises due to the vast number of possible content combinations (i.e., total contents choose cache size). Our future research will extend this work by incorporating energy planning through simulations that consider UAV flight dynamics, communication-based depletion, and other energy-consuming factors.

From an algorithmic viewpoint, research in [13] proposes optimizations in flight trajectory, communication scheduling, service coverage through optimized hovering time, and multi-hop relaying across multiple UAVs. In another approach, authors in [14] target similar objectives for IoT networks, utilizing multi-hop device-to-device (D2D) routing to extend coverage for energy-constrained UAVs. Although these studies address coverage extension, they do not engage with content placement and caching issues, which are central to our work here. The issue of content placement and caching is tackled in [15]-[18]. [15] proposes using named data networking (NDN) architecture within IoT networks, where UAVs collect and deliver data to interested recipients to prevent retransmissions. In [16], UAVs proactively transmit content to a selected subset of ground nodes that cooperatively cache necessary contents. Another paper [17] introduces a probabilistic cache placement technique aimed at maximizing cache hit probabilities in networks with wireless nodes organized via a homogeneous Poisson Point Process. Additionally, [18] addresses denial-of-service challenges associated with UAVs in communication, though without considering storage constraints, which are pertinent to the problem discussed in this paper.

To further the research on algorithmic methods, studies have explored traffic offloading and learning-based caching strategies. For instance, [19] discusses enhancing the effective caching capacity of UAVs by accounting for the popularity and size of stored content. Research in [20] proposes a UAV-enabled small-cell network where UAVs offload data traffic from small-cell base stations (SBSs), that caches the most popular content to directly serve user demands as required. Similarly, [21] attempts to reduce ground base-station traffic load through UAV caching. In this approach, a joint caching and UAV trajectory optimization method using particle swarm optimization treats each caching strategy as a particle [22]. Another study [23] develops a strategy to minimize content delivery delays by jointly optimizing UAV trajectories and radio resource allocation, using a deep Q-learning approach to handle optimization challenges in large networks with extensive state-action spaces. However, while these methods address caching decisions using traffic offloading and learning-based strategies, they often overlook content popularity heterogeneity, which limits their effectiveness in scenarios where content demand varies by user location.

Another algorithmic advancement is presented in [24], where UAV trajectory control mechanisms determine whether to continue serving users along a trajectory or to return to the charging station based on real-time conditions. In a similar study, [25] employs a reinforcement learning-based UAV-caching decision-making model, using content requests, storage availability, and buffer states as inputs for a Markov Decision Process. However, these studies do not examine the impacts of disaster geography, demand variability, and UAV trajectory effects on caching policy, which are central to this paper. Scalability and robustness are critical in UAV-assisted caching, particularly in high-mobility and dense environments. The work in [43] explores reinforcement learning for UAV trajectory optimization to maintain network coverage in dynamic rescue scenarios. While their focus is on trajectory control, the approach in this paper is trajectory-agnostic, meaning it can complement any UAV mobility strategy without imposing specific movement constraints. FedMAB enhances scalability by optimizing content selection in a decentralized manner which reduces dependency on UAV repositioning. Given the non-aligned objectives of UAV trajectory optimization and caching-based adaptability, direct comparisons require careful contextual alignment. While FedMAB is inherently designed to operate effectively in large-scale deployments, future evaluations will further assess its performance under extreme mobility and node density to reinforce its adaptability.

While some UAV-based caching methods [20, 21] can partially support scenarios with partial infrastructure failure, they are less effective when backhaul is entirely destroyed, and a fully functional alternative is needed. Furthermore, most of these methods [19]-[21] assume static global content popularity [26], thereby missing the nuances of real-world demand variability and time-sensitive content needs in disaster scenarios. Optimization methods in [22]-[25] rely on long-term estimates, which inherently lack adaptability to dynamic network conditions and fluctuating demand. Additionally, prior research lacks explicit strategies for maximizing cache space and minimizing costly downloads through vertical links.

The integration of UAV-assisted content caching and optimization has also been explored in vehicular networks. The work in [44] presents a dynamic UAV-assisted data sharing framework leveraging an aerial Peer-to-Peer (P2P) backbone. This approach optimizes UAV trajectories using a Deep Q Network (DQN) to enhance lookup success rates and minimize data retrieval latency. Our proposed Federated Multi-Armed Bandit learning method similarly aims to improve content availability, but focuses on distributed learning for adaptive caching in disaster-affected environments. The insights from 44] reinforce the importance of adaptive learning mechanisms for UAV-assisted data dissemination and validate the efficacy of reinforcement learning techniques and its variants like Multi-Armed Bandits in such applications.

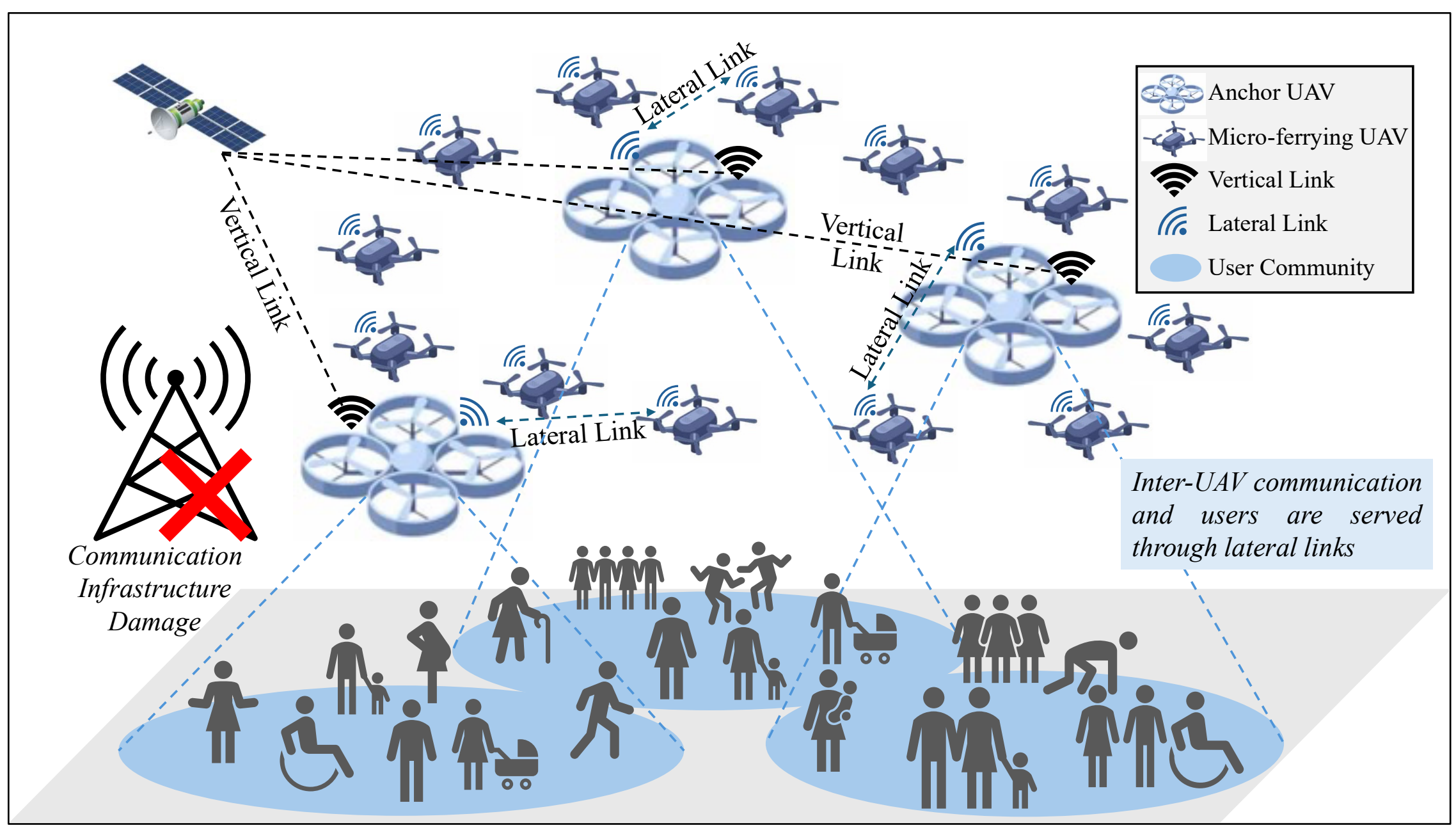

Fig. 1. Coordinated UAV system for content dissemination in environments without communication infrastructure.

To address these challenges, this paper introduces a Federated Multi-Armed Bandit learning model for UAV caching decisions that considers geo-temporal content popularity variations and demand heterogeneity. The model also employs selective caching algorithms to enhance system performance by sharing information between anchor UAVs and micro-ferrying UAVs [27]. This approach targets adaptability, cache space maximization, and reducing costly server downloads via vertical links, factors that have not been comprehensively addressed in prior UAV-caching research [13]-[25].

## 3 SYSTEM MODEL

### 3.1 UAV Hierarchy

As shown in Fig. 1, a two-tiered UAV-assisted content dissemination system is deployed. Each community is served by a dedicated A-UAV, which operate with much larger power budgets compared to Micro-UAVs described next. The A-UAVs use lateral wireless connections (i.e., WiFi etc.) to communicate with users in that community. A-UAVs can download content via an expensive vertical link such as satellite-based internet. The system introduces a set of low-power-budget Micro-UAVs [29] for the role of ferrying (MF-UAVs). MF-UAVs are mobile and possesses only lateral communication links such as Wi-Fi. Unlike the A-UAVs, the MF-UAVs do not possess expensive vertical communication interfaces such as satellite links etc. Effectively, the MF-UAVs act as content transfer agents across different user communities by selectively caching and transferring content across the A-UAVs through their lateral links.

*Discussion*: The concept of hierarchical UAV structuring and content clustering is aligned with the principles of efficient content dissemination. While the proposed framework does not explicitly impose a higher-layer structure for clustering communities, aspects of hierarchical coordination already emerge through the two-tier UAV system. The Anchor UAVs (A-UAVs) inherently serve as local caching coordinators for their respective communities, while Micro-Ferrying UAVs (MF-UAVs) transport content across different regions, effectively creating a layered distribution system without rigid structuring.

Additionally, content placement decisions in FedMAB, to be discussed later, naturally result in implicit clustering, as the learning model prioritizes frequently requested content within specific regions which ensures that communities receive relevant cached data without requiring manual segmentation. This data-driven approach allows the system to dynamically adapt to evolving content access patterns rather than relying on predefined clusters.

The model also integrates an implicit indexing mechanism through its Q-value system, also to be discussed in the forthcoming sections, where content importance is dynamically ranked based on request patterns. This ranking ensures that MF-UAVs retrieve and distribute the most relevant content without needing a predefined indexing structure.

### 3.2 Content Demand and Provisioning Model

The content popularity distribution, quality of services and content provisioning are outlined below.

Content Popularity: Research has shown that user content request patterns often follow a power law distribution such as the Zipf distribution [28]. In Zipf distribution, the popularity of a content is proportional to the inverse of its rank, and is a geometric multiple of the next popular content. Popularity of content '$i$' is given as:

$$\mathcal{P}_{\alpha}(i) = \left(\frac{1}{i}\right)^{\alpha} \Big/ \sum_{k \in C} \left(\frac{1}{k}\right)^{\alpha} \tag{1}$$

The Zipf parameter $\alpha$ determines the distribution's skewness, while the total number of contents in the pool is represented by the parameter $C$. The inter-request time from a user follows the popular exponential distribution [28].

Note that Zipf distribution is widely recognized as an appropriate model for content popularity, with empirical validation across multiple domains, including publications, online video platforms, social media, and recommendation systems such as Netflix, Instagram and more. This distribution effectively captures the heavy-tailed nature of content requests, where a small fraction of items accounts for the majority of demand.

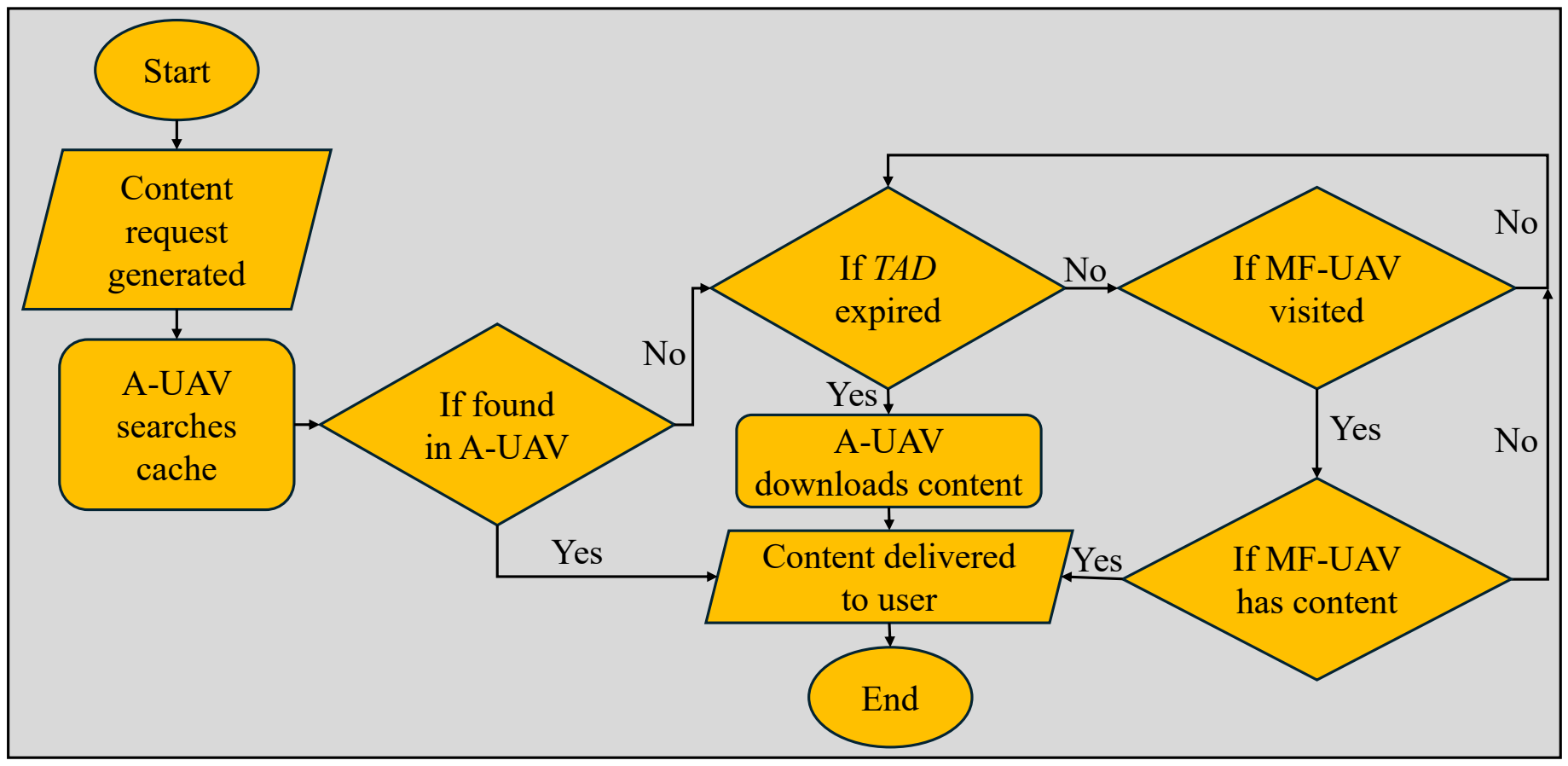

Fig. 2. Content Delivery Process

Tolerable Access Delay (*TAD*): For each generated request, a *TAD* [5] is specified. *TAD* is a Quality-of-Service parameter that indicates the duration that a user is ready to tolerate before its requested content must be provisioned. Operationally, if a content is not available from the UAVs within the specified *TAD*, it must be downloaded from a central server using the expensive vertical links of A-UAVs. To be noted that $TAD$ is request specific in which it is different for different contents depending on the requesting user's urgency.

Content Provisioning: Upon receiving a request from one of its community users, the locally deployed A-UAV first searches its local storage for the content. If the content is not found, the A-UAV waits for a potential future delivery by a traveling MF-UAV. If no MF-UAV arrives with the requested content within the specified *TAD*, the A-UAV then proceeds to download it through its vertical link. Since vertical links such as satellite links are expensive, smart caching strategies that can make the content accessible from the UAVs can be effective in reducing the overall content provisioning costs.

## 4 CONTENT CACHING PROBLEM FORMULATION

The caching problem focuses on selecting which contents should be stored at UAVs to maximize content availability while considering storage constraints, access latency, and varying user demand across communities. For a given number of Anchor and Micro-ferrying UAVs, the caching problem at the UAVs can be defined as follows.

$$\underset{\forall n \in \mathcal{N}}{maximize} \left[ \frac{1}{\mathcal{N}} \left[ \sum_{n=1}^{\mathcal{N}} \mathbb{P}_n^{avail} \right] \right] \tag{2}$$

$$such\ that \sum_{i=1}^{N_A} |C_A| + \sum_{j=1}^{N_{MF}} |C_{MF}| < |C| \tag{3}$$

$$and\ \mathcal{T}_{\hbar}^{serve} - \mathcal{T}_{\hbar}^{req} \leq TAD_{\hbar}, \hbar \in \mathcal{H}_n, \forall \mathcal{H}_n = \{1,2,3,\cdots\} \tag{4}$$

where, $\mathbb{P}_n^{avail} = \frac{\mathcal{H}_n}{\mathcal{R}_n}$, $\mathcal{H}_n$ is the number of contents provisioned at community '$n$' by the UAV system (both A-UAVs and MF-UAVs), $\mathcal{R}_n$ is the total number of requests made by users at community '$n$', $\mathcal{N}$ is the number of commmunities, $C$ is the total contents in pool, $C_A$ is the cache of each A-UAV, $C_{MF}$ is the cache of MF-UAVs, $N_A$ is the number of A-UAVs, $N_{MF}$ is the number of MF-UAVs, $\mathcal{T}_{\hbar}^{req}$ is the time at which a content '$\hbar$' is requested by a user, $\mathcal{T}_{\hbar}^{serve}$ is the time when content '$\hbar$' is served to the user by the UAV system, and $TAD_{\hbar}$ is the tolerable access delay associated with content '$\hbar$'. The caching problem focuses on maximizing the overall content availability, as shown in Equation. 2. This objective is constrained by maintaining the cumulative caching capacities of the UAVs below the total number of contents in the content pool, which is captured in Equation. 3. An additional constraint is imposed by the tolerable access delay associated with a content served to the user by the UAV-aided system (refer to Equation. 4).

## 5 BENCHMARK CACHING POLICY WITH A-PRIORI DEMAND KNOWLEDGE

This section focuses on the following caching related design questions: a) which content to be downloaded and cached in the A-UAVs so that they can serve their own community directly, and the remote communities via the traveling MF-UAVs; b) which contents to be cached when the popularity and $TAD$ of contents vary at different communities; c) which content to be transferred from the A-UAVs to the MF-UAVs; and, d) what is the benchmark caching policy with heterogeneous content popularity at each user community and heterogeneity in request-specific $TAD$.

These questions are addressed by formulating a benchmark caching policy with *a priori* known heterogeneous content popularities. This benchmark caching policy also considers and modifies the caching policy to cater to the request specific $TAD$s. After understanding the benchmark, runtime and dynamic mechanisms will be developed in a next section.

### 5.1 Caching at Anchor UAVs (A-UAVs)

For simplicity, let us consider a disaster/war-stricken area with homogeneous content popularity across all the user communities. An A-UAV is assigned to each community for content provisioning. The number of A-UAVs in the system is denoted by $N_A$. In such a scenario, the effective caching capacity of A-UAVs can be maximized by storing a certain number of unique contents in all the A-UAVs, and share those contents across the communities via the traveling MF-UAVs. To maximize the effective caching capabilities of all $N_A$ A-UAVs, the cache space of each A-UAV is divided into two segments [28], namely, *Segment-1* and *Segment-2*. Let the sizes of Segment-1 and Segment-2 of the A-UAV cache be $|C_{S1}|$ and $|C_{S2}|$ respectively. They can be expressed as follows:

$$|C_{S1}| = \lambda \times |C_A| \tag{5}$$

$$|C_{S2}| = (1-\lambda) \times |C_A| \tag{6}$$

where $\lambda$ is a *storage segmentation factor* (SSF) that decides the split between the segments within a A-UAV [28]. The top $\lambda.|C_A|$ popular contents are cached in Segment-1. These contents are same across all A-UAVs whereas contents stored in Segment-2 are different. This results into the number of total Segment-2 contents stored across all $N_A$ A-UAVs to be:

$$|C_{S2}^{total}| = N_A \times (1-\lambda) \times |C_A| \tag{7}$$

These contents are shared across all user communities via the mobile MF-UAVs. These contents have popularities after the top $\lambda.|C_A|$ popular Segment-1 contents in all the A-UAVs. For symmetry, all $N_A \times (1-\lambda) \times |C_A|$ Segment-2 contents are uniformly randomly distributed across $N_A$ number of A-UAVs. Hence, for a given Zipf parameter $\alpha$ which determines the distribution's skewness, the total number of contents in the system is as follows:

$$|C_{sys}^{\alpha}| = \lambda \times |C_A| + N_A \times (1-\lambda) \times |C_A| \Rightarrow |C_{sys}^{\alpha}| = \left(\lambda + N_A \times (1-\lambda)\right) \times |C_A| \tag{8}$$

Now consider a heterogeneous demand scenario in which every community has a different demand pattern, and each content is requested with a fixed pre-decided $TAD$. The above caching policy is modified as follows to address such a situation. Some contents from Segment-1, termed as *exclusive* contents, are cached in one or some of the A-UAVs, but not in all of them [26]. Whereas the remaining contents from Segment-1, termed as *non-exclusive* contents, are cached at all the A-UAVs [28]. Therefore, unlike the homogeneous popularity scenario, the number of contents in Segment-1 across all A-UAVs may be more than $\lambda \times |C_A|$ due to the different A-UAV specific exclusive contents. This shown below:

$$|C_{S1}^{total}| = |C_{NE}| + |C_E^{total}| \geq \lambda \times |C_A| \tag{9}$$

Similar to the caching policy in a homogeneous popularity scenario, contents in Segment-2 do not repeat across the A-UAVs. If $C_{NE}$ $and$ $C_E^{total}$ are the non-exclusive and total exclusive contents in Segment 1, then total number of contents in the system can be modified from Equation. 8, and can be expressed as follows:

$$|C_{sys}^{\alpha}| = |C_{NE}| + |C_E^{total}| + N_A \times (1-\lambda) \times |C_A| \Rightarrow |C_{sys}^{\alpha}| \geq \left(\lambda + N_A \times (1-\lambda)\right) \times |C_A| \tag{10}$$

The classification of content as exclusive or non-exclusive is determined by predefined access constraints that specify whether a piece of content is intended for a single community or multiple communities. Exclusive content is assigned to a specific segment and is cached at designated A-UAVs serving that community. Non-exclusive content is intended for broader dissemination and is made available across multiple A-UAVs to maximize accessibility. In the benchmark models, these classifications dictate where content is stored which ensures exclusive content remains within its intended segment while non-exclusive content is widely distributed. However, in FedMAB, that is to be discussed in the forthcoming section, caching decisions are not constrained by predefined exclusivity labels. Instead, the learning model determines caching locations dynamically based on observed request patterns. Content initially classified as exclusive or non-exclusive may be placed in different locations if the bandit-driven learning process identifies a more efficient caching strategy. This ensures that caching adapts to real-world demand rather than being restricted by static classifications.

To be noted that the above stated caching policies take the contents' popularity into consideration while making the caching decisions. However, the promptness with which a content needs to be provisioned, i.e., the $TAD$, may not always be positively correlated with its popularity. Therefore, unlike cache space optimization done till now, the caching policy needs modification from a perspective that considers a content's importance. Hence, unlike the cache space optimization undertaken thus far, the caching policy requires modification from a standpoint that considers the significance of content.

Now consider a demand heterogeneous scenario where every community has a different demand pattern, and each content is requested with its own specific $TAD$ [26]. If a content is requested with less $TAD$, this implies that the user is not willing to wait for a visiting MF-UAV to deliver the content. Therefore, caching such time-critical contents at the A-UAVs becomes imperative. To prioritize caching of such contents in Segment-1 of A-UAVs, this paper devices a value-based caching policy where the value of a requested content '$\hbar$' is calculated from its popularity and its $TAD$, and is as follows:

$$\mathcal{V}(\hbar) = \kappa \times \frac{TAD_{min}}{p_{\alpha}(1)} \times \frac{\mathcal{P}_{\alpha}(\hbar)}{TAD_{\hbar}} \Rightarrow \mathcal{V}(\hbar) = \kappa \upsilon \times \frac{\mathcal{P}_{\alpha}(\hbar)}{TAD_{\hbar}} \tag{11}$$

Here, $\mathcal{P}_{\alpha}(\hbar)$ is the popularity of the content as per Zipf Distribution, $TAD_{\hbar}$ is the tolerable access delay associated with the content request, $\kappa \in [0,1]$ is a scalar weight which increases with decrease in popularity and $\upsilon$ is a *normalization constant*. For a given Zipf (popularity) parameter $\alpha$, the normalization constant is calculated from the minimum possible $TAD$ ($TAD_{min}$) and the maximum possible popularity, which is $\mathcal{P}_{\alpha}(1)$. The quantity $\mathcal{V}(\hbar)$ is bounded between $[0, 1]$, it increases with increase in $\mathcal{P}_{\alpha}(\hbar)$, and it decreases with $TAD_{\hbar}$. This value-based caching policy increases the likelihood of contents requested with low $TAD$ to be cached *in Segment-1* of the A-UAVs, thus making them more readily available. To be noted that the cache space maximization method developed in Equations. 5-10 still applies to this scenario. Here, the contents to be cached are chosen based on their values instead of their popularity, which is shown below:

$$\left|C_{sys}^{\mathcal{V}}\right| = |C_{NE}| + \left|C_{E}^{total}\right| + N_A \times (1-\lambda) \times |C_A| \tag{12}$$

### 5.2 Caching at Micro-Ferrying UAVs (MF-UAVs)

The purpose of the MF-UAVs is to ferry $\left|C_{E}^{total}\right| + N_A \times (1-\lambda) \times |C_A|$ number of contents stored across $N_A$ number of A-UAVs (see Equation. 10). Due to the limitations of per-MF-UAV caching space [29] (i.e., $|C_{MF}|$), their caching policy should be determined based on the trajectories, the value of $\lambda$, the Zipf popularity, and the $TADs$ associated with the contents to be cached [30].

Consider a situation in which an MF-UAV *'$j$'* is approaching towards the A-UAV *'$i$'*. Let $U_i$ be the set of all exclusive contents in Segment-1 of all A-UAVs and all contents from Segment-2 of all A-UAVs in the entire system except the ones stored in A-UAV *'$i$'*. To maximize content availability for the users in A-UAV *$i$'s* community, the MF-UAV should carry $|C_{MF}|$ top valued contents (refer to Equation. 12) from the set $U_i$ while approaching A-UAV *i*.

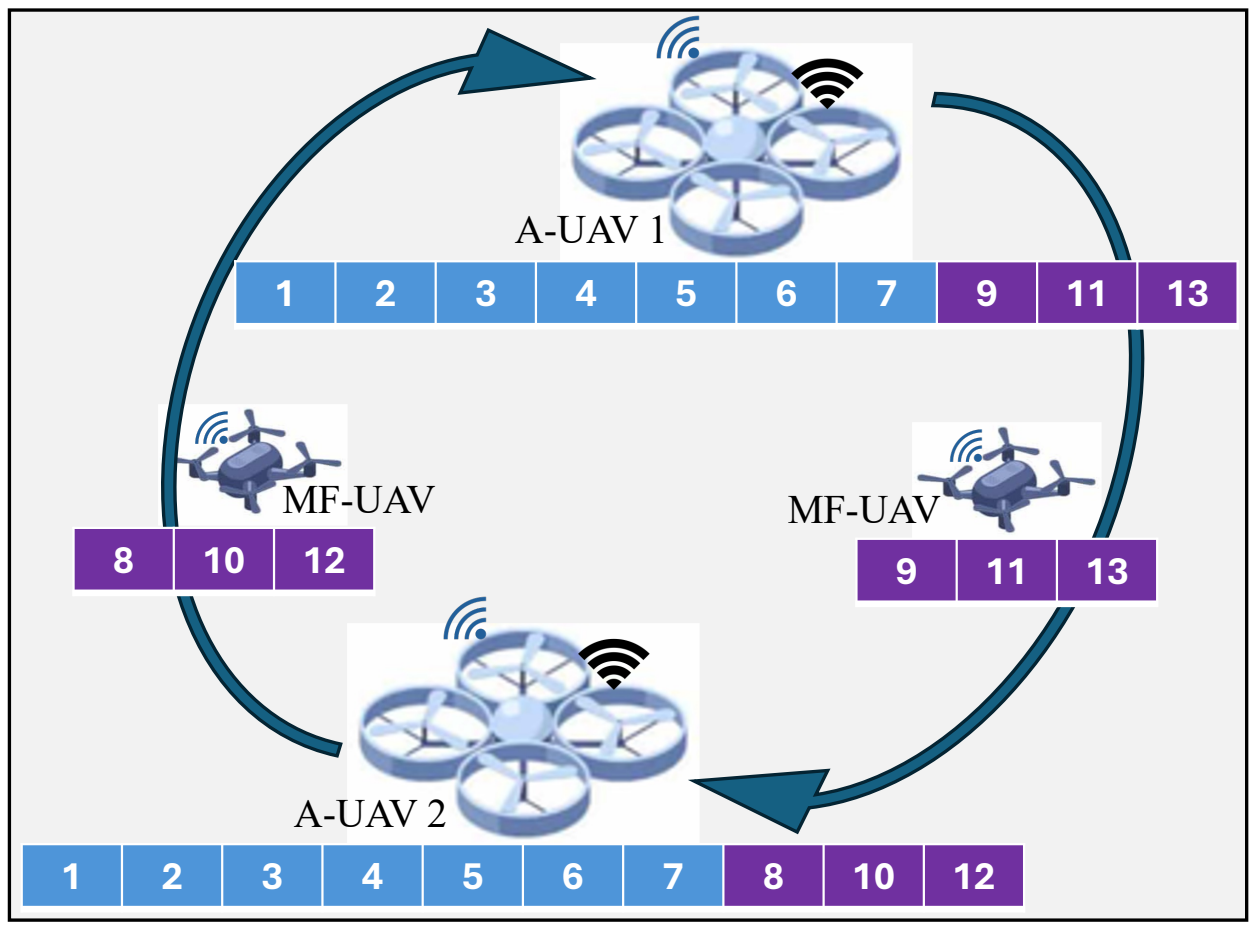

Fig. 3. Caching Policy at MF-UAVs

The size of the set $U_i$ can be expressed as $|U_i| = \left|C_E^{total}\right| + (N_A - 1).(1 - \lambda).|C_A|$. In scenarios when $|C_{MF}| \leq |U_i|$, the MF-UAV should carry the $|C_{MF}|$ top popular contents as outlined above. Otherwise, the MF-UAV should carry all $|U_i|$ contents, leaving part of the MF-UAV cache (i.e., $|C_F| - |U_i|$) empty. This implies that an apt choice of caching policy at A-UAVs affect the utilization of MF-UAV's cache.

MF-UAV caching policy is explained in the pseudocode below.

**Algorithm 1.** MF-UAV Caching Algorithm with Value-based policy executed at the A-UAVs

1. **Input:** Total A-UAVs in its trajectory, $TAD$, next A-UAV '$i$', present A-UAV '$i-1$'
2. **Output:** $C_{MF}$ contents for MF-UAV '$j$'
3. **Initialize** $C_A$ contents in each A-UAV based on value of contents
4. **while** True:
5.     **if** M**F-UAV leaving for next A-UAV '$i$' **then do**
6.         **for** $k = 0$ $to$ $length$(A-UAV '$i$' cache $C_A^i$) **do**
7.             Check if $k$ in $C_{MF}$ cache of MF-UAV '$j$'
8.             **if** true **then do**
9.               Replace '$k$' with highest value content from $C_A^{i-1}$ not cached in MF-UAV '$j$' & A-UAV '$i$'
10.             **end if**
11.         **end for**
12.     **end if**
13.     **Update** next A-UAV '$i$', present A-UAV '$i-1$'
14. **end while**

### 5.3 Theoretical Performance Upper-Bound

In this section, a theoretical performance upper-bound is computed when the A-UAVs and MF-UAVs follow the benchmark caching policy as described in Section 5.1. Let us consider a UAV-caching system where there are $N_A$ number of A-UAVs, and $N_{MF}$ number of MF-UAVs. The number of MF-UAVs traveling in a group is denoted by $N_{MF}^G$. MF-UAVs traverse the complete disaster region in $\mathcal{T}_{Cycle}$ seconds. The hover ratio is $\mathcal{R}_{Hov}$, which is the ratio of the time an MF-UAV stays at a community before leaving for the next to $\mathcal{T}_{Cycle}$. The transition ratio is $\mathcal{R}_{Trans}$, which is the ratio between the time on MF-UAV takes to travel from one community to the next and $\mathcal{T}_{Cycle}$. For simplicity, the inter-community distances are kept the same. The content request pattern is heterogeneous across communities with popularity parameter of $\alpha$. Every request '$\hbar$' is accompanied by its respective $TAD_{\hbar}$.

The performance upper-bound has three important parts, namely, the probability $\mathbb{P}_A$ that the content is found in an A-UAV '$i$', the probability of a content being found in MF-UAV $\mathbb{P}_{MF}$, and the probability that an MF-UAV is accessible near a A-UAV before content requests expire $\mathbb{P}_{Access}$. The accessibility probability $\mathbb{P}_{Access}$ is computed according to a condition $\mathbb{T}_{cond}$ which is given below:

$$\mathbb{T}_{cond} = \left(\frac{N_{MF}^{G} \times N_A}{N_{MF}} - 1\right) \times \mathcal{R}_{Hov} \times \mathcal{T}_{Cycle} + \left(\frac{N_{MF}^{G} \times N_A}{N_{MF}}\right) \times \mathcal{R}_{Trans} \times \mathcal{T}_{Cycle}$$

$$\Rightarrow \left(\left(\frac{N_{MF}^{G} \times N_A}{N_{MF}} - 1\right) \times \mathcal{R}_{Hov} + \left(\frac{N_{MF}^{G} \times N_A}{N_{MF}}\right) \times \mathcal{R}_{Trans}\right) \times \mathcal{T}_{Cycle} \qquad (13)$$

Equation. 13 computes the time an MF-UAV takes to revisit a location. To ensure that the formulation remains applicable across different UAV configurations, it is important to clarify how the grouping of MF-UAVs and the accessibility factors influence the caching process. The first term in parentheses in Equation. 13 does not become zero, because $N_{MF}^{G}$ represents a dynamically determined grouping of MF-UAVs based on their flight dynamics and proximity. Since the grouping varies depending on how MF-UAVs ferry content and reduce redundant transmissions, the ratio $\frac{N_{MF}^{G}.N_A}{N_{MF}}$ does not equal one which ensures the first term remains nonzero. Consequently, the first term does not become negative because all involved parameters are strictly positive, and by definition, $N_{MF}^{G} \leq N_{MF}$. The subtraction of one ensures that the formulation correctly accounts for accessibility relative to the number of A-UAVs in the MF-UAVs' trajectory cycle.

Additionally, $\mathcal{R}_{Hov}$ is formulated as a probability weight rather than a strict ratio to $\mathcal{T}_{Cycle}$ that ensures adaptability in determining the weighted contribution of hovering time. By treating $\mathcal{R}_{Hov}$ as a probability weight, the framework allows for dynamic adjustments based on MF-UAV group behavior which prevents an over-simplified proportionality that does not account for variations in flight patterns and spatial arrangements. This ensures that the influence of hovering time is contextually adjusted rather than statically imposed, preserving the generality of the formulation.

Depending on the condition being satisfied, $\mathbb{P}_{access}$ is computed using the following piece-wise expression:

$$\mathbb{P}_{access} = \begin{cases} \dfrac{N_{MF} \times \left(\mathcal{R}_{Hov}\mathcal{T}_{Cycle} + \overline{TAD}\right)}{N_{MF}^{G} \times N_A \times \left(\mathcal{R}_{Hov}\mathcal{T}_{Cycle} + \mathcal{R}_{Trans}\mathcal{T}_{Cycle}\right)}, & for\ \overline{TAD} < \mathbb{T}_{cond} \\ 1, & for\ \overline{TAD} \geq \mathbb{T}_{cond} \end{cases}$$

$$= \begin{cases} \dfrac{N_{MF} \times \left(\mathcal{R}_{Hov}\mathcal{T}_{Cycle} + \overline{TAD}\right)}{N_{MF}^{G} \times N_A \times \left((\mathcal{R}_{Hov} + \mathcal{R}_{Trans})\mathcal{T}_{Cycle}\right)}, & for\ \overline{TAD} < \mathbb{T}_{cond} \\ 1, & for\ \overline{TAD} \geq \mathbb{T}_{cond} \end{cases} \qquad (14)$$

Here, $\overline{TAD}$ is the mean $TAD$, which is used for generalization. The second part of the piece-wise expression in Equation. 14 shows that for a very large $TAD$, the contents in an MF-UAV are always accessible. However, for $\overline{TAD}$ less than the $\mathbb{T}_{cond}$, the contents in MF-UAVs are partially accessible. Note that the physical accessibility to MF-UAVs does not guarantee the access to a requested content since the MF-UAVs can store only a limited number of contents. The probability $\mathbb{P}_{MF}$ that a content can be found in a MF-UAV is given below:

$$\mathbb{P}_{MF} = \left[ \sum_{\hbar \in \{|C_E^{total}| + N_A.(1-\lambda).|C_A|\}}^{\hbar \notin \{|C_{NE}^i| + |C_E^i|\}} \mathcal{V}(\hbar) \right] \Big/ \left[ \sum_{\forall C} \mathcal{V}(\hbar) \right] \Rightarrow \mathbb{P}_{MF} = \frac{\sum_{\hbar \in \{|C_E^{total}| + N_A.(1-\lambda).|C_A|\}}^{\hbar \notin \{|C_{NE}^i| + |C_E^i|\}} \kappa \times \frac{TAD_{min}}{p_\alpha(1)} \times \frac{\mathcal{P}_\alpha(\hbar)}{TAD_\hbar}}{\sum_{\hbar=1}^{C} \kappa \times \frac{TAD_{min}}{p_\alpha(1)} \times \frac{\mathcal{P}_\alpha(\hbar)}{TAD_\hbar}} \tag{15}$$

The above expression considers the value of the contents from Equation. 11. Now, $\mathbb{P}_A$, the probability of finding a requested content in the local A-UAV of the request generating community, is expressed as:

$$\mathbb{P}_A = \left[ \sum_{\forall |C_{NE}| + |C_E|} \mathcal{V}(\hbar) \right] \Big/ \left[ \sum_{\forall C} \mathcal{V}(\hbar) \right] \Rightarrow \mathbb{P}_A = \left[ \sum_{\hbar \in |C_{NE}| + |C_E|} \kappa \upsilon \times \frac{\mathcal{P}_\alpha(\hbar)}{TAD_\hbar} \right] \Big/ \left[ \sum_{\hbar=1}^{C} \kappa \upsilon \times \frac{\mathcal{P}_\alpha(\hbar)}{TAD_\hbar} \right] \tag{16}$$

Combining Equations. 14, 15 and 16, the average content availability at a community '$n$' can be expressed as:

$$\mathbb{P}_n^{avail} = \mathbb{P}_A + \mathbb{P}_{access} \times \mathbb{P}_{MF} \tag{17}$$

Equation. 17 shows that the contents from A-UAV '$i$' and contents from future visiting MF-UAVs contribute towards the average availability $\mathbb{P}_n^{avail}$ at community '$n$' within the specified $TADs$. Note that all unavailable contents within specified $TADs$ will be downloaded by the A-UAVs using their expensive vertical links such as a Satellite Internet link. Thus, availability indirectly indicates the content download cost in the system.

The aim of the learning-based caching policy, discussed in the next section, is to achieve the above-mentioned benchmark performance in terms of content availability. The proposed learning is achieved in a distributed manner in which all UAVs learn the caching policy without *a priori* demand information and without explicit sharing of user request data.

## 6 FEDERATED MULTI-ARMED BANDIT LEARNING FOR CONTENT CACHING

### 6.1 Caching Policy using Top-k Multi-Armed Bandit

Upon deployment in a community, a A-UAV's primary task is to optimize content availability for users by determining which contents to download and cache through its vertical link. One of the ways to approach this objective involves the utilization of a *Top-k* Multi-Armed Bandit (*Top-k* MAB) learning agent within the A-UAV.

The *Top-k* MAB learning, a variant of the classical Multi-Armed Bandit problem in reinforcement learning, is employed to maximize the cumulative reward $\mathbb{R}(T)$ over a finite time horizon $T$ [31]. In contrast to the traditional MAB, this variant involves choosing $k$ arms simultaneously from a set of $M$ arms and receiving individual rewards for each arm selected.

$$\mathbb{R}_T = \max \left[ \sum_{t=1}^{T} \left( \sum_{i=1}^{k} \mathbb{E}[\mathbb{R}_t(i)] \right) \right] \tag{18}$$

Each A-UAV is assumed to be equipped with a *Top-k* MAB agent. Here, the selection of content for caching corresponds to choosing an arm, with '$k$' in '*Top-k*' representing the caching capacity ($C_A$) of the A-UAV. The agent's objective is to choose '$C_A$' contents from a larger set of '$C$' contents in order to maximize content availability for users.

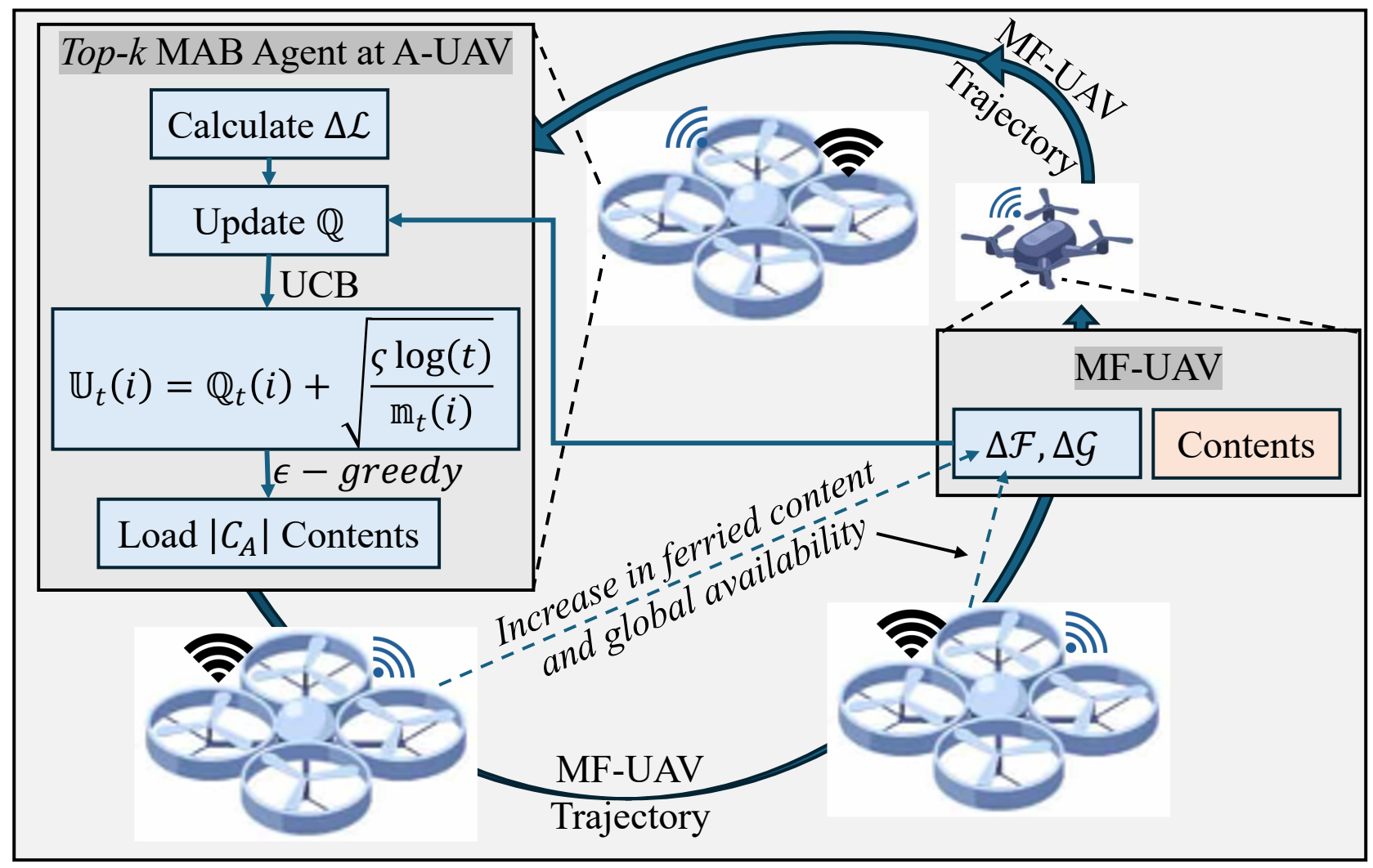

Fig. 4. *Top-k* Multi-Armed Bandit Learning for Caching Policy at A-UAVs

In the UAV-aided content dissemination environment, A-UAVs interact by selecting specific content sets (i.e., MAB actions) for caching. The feedback from the environment for the taken actions are in the form of rewards/penalties. Micro-ferrying UAVs play a vital role in transferring information across the system, contributing to the computation of rewards and penalties. Actions are rewarded when cached contents are requested and served within the tolerable access delay. Otherwise, they are penalized.

The learning epoch for each *Top-k* MAB agent is strategically chosen based on the MF-UAVs' accessibility at the corresponding community. Therefore, epoch duration is influenced by the visiting frequency of MF-UAVs. MF-UAVs carry the content availability information of the already visited A-UAVs in its trajectory. The *Top-k* MAB agents leverage such information and learn to cache contents through a multi-dimensional reward structure, encompassing the *local*, *ferrying*, and *global* rewards. These rewards are contingent upon the availability of the sets of locally served contents ($\mathcal{L}$), contents served at other communities via ferrying ($\mathcal{F}$), and overall contents served across all communities ($\mathcal{G}$). These contents can be served to the users directly by a A-UAV or indirectly via the visiting MF-UAVs. If a cached content is served to a user within the given *TAD*, and an increase in content availability is observed, the caching decision for the content is rewarded. The type of reward is determined by the set to which the cached content belongs to. The expressions for three types of rewards are given as follows:

$$\mathbb{R}(i,\mathcal{L}) = [(i \in \mathcal{L}) \wedge (\Delta_{\mathcal{L}} \geq 0)] - [(i \notin \mathcal{L}) \wedge (\Delta_{\mathcal{L}} < 0)] \tag{19}$$

$$\mathbb{R}(i,\mathcal{F}) = \frac{1}{N_A - 1} \sum_{j=1, j \neq \mathbb{X}}^{N_A} [(i \in \mathcal{F}) \wedge (\Delta_{\mathcal{F}} \geq 0)] - \frac{1}{N_A - 1} \sum_{j=1, j \neq \mathbb{X}}^{N_A} [(i \notin \mathcal{F}) \wedge (\Delta_{\mathcal{F}} < 0)]$$

$$\Rightarrow \mathbb{R}(i,\mathcal{F}) = \frac{1}{N_A - 1} \sum_{j=1, j \neq \mathbb{X}}^{N_A} [(i \in \mathcal{F}) \wedge (\Delta_{\mathcal{F}} \geq 0)] + \frac{1}{N_A - 1} \sum_{j=1, j \neq \mathbb{X}}^{N_A} \Big[ [\neg [(i \notin \mathcal{F}) \wedge (\Delta_{\mathcal{F}} < 0)]] - 1 \Big] \tag{20}$$

$$\mathbb{R}(i,\mathcal{G}) = \frac{1}{N_A}\sum_{j=1}^{N_A}\left[(i \in \mathcal{G}) \wedge \left(\Delta_{\mathcal{G}} \geq 0\right)\right] - \frac{1}{N_A}\sum_{j=1}^{N_A}\left[(i \notin \mathcal{G}) \wedge \left(\Delta_{\mathcal{G}} < 0\right)\right]$$

$$\Rightarrow \mathbb{R}(i,\mathcal{G}) = \frac{1}{N_A}\sum_{j=1}^{N_A}\left[(i \in \mathcal{G}) \wedge \left(\Delta_{\mathcal{G}} \geq 0\right)\right] + \frac{1}{N_A}\sum_{j=1}^{N_A}\Big[\left[\neg\left[(i \notin \mathcal{G}) \wedge \left(\Delta_{\mathcal{G}} < 0\right)\right]\right] - 1\Big] \quad (21)$$

$$where, [A] = \begin{cases} 1, & if\ A\ is\ true \\ 0, & otherwise \end{cases}$$

The above equations are used for computing the reward received by a *Top-k* MAB agent at A-UAV '$\mathbb{X}$'. Caching content '$i$' at A-UAV '$\mathbb{X}$' is rewarded if it leads to an increase in availability. Here, $\mathbb{R}(i,\mathcal{L})$, $\mathbb{R}(i,\mathcal{F})$, and $\mathbb{R}(i,\mathcal{G})$ are local, ferrying and global rewards, respectively. The terms $\Delta_{\mathcal{L}}$, $\Delta_{\mathcal{F}}$ and $\Delta_{\mathcal{G}}$ correspond to the increase in local availability, ferried content availability, and global availability, respectively. Each type of reward is contingent upon satisfying the condition '$f(i)$' in the *Iverson bracket* "$[f(i)]$". The first terms in Equations. 7, 8 and 9 represent the reward accumulated by caching content '$i$' cached at A-UAV '$\mathbb{X}$', whereas the second term is the penalty associated with adverse condition. To be noted that $\mathbb{R}(i,\mathcal{F})$, and $\mathbb{R}(i,\mathcal{G})$ are higher if the content '$i$' is requested and served at more number of communities.

Learning employs a tabular approach where a Q-table is maintained for all contents in A-UAVs. Each content corresponds to a Q-value or action-value [3] in the Q-table. The Q-value indicates the importance of a content depending on its popularity and frequency of request. Additionally, it indirectly captures the geographical relevance of the content which is related to where the content has been requested in the disaster region. The *Top-k* MAB agent updates the Q-value for a content at each learning epoch based on the multi-dimensional rewards (Equations 19-21). These rewards are derived from the interactions of a A-UAV's agent with the UAV-aided content dissemination system, shaping its understanding of optimal actions (contents to cache). The recursive Q-value update expression for content '$i$' at A-UAV "$\mathbb{X}$" is given as follows:

$$\mathbb{Q}_{t+1}(i) = (1-\alpha_L)\mathbb{Q}_t(i) + \alpha\left\{\mathbb{R}_t(i,\mathcal{L}) + \begin{pmatrix} [(x,y,z)_{A-UAV'\mathbb{X}'} = (x,y,z)_{MF-UAV}] \\ \times\left(\mathbb{R}_t(i,\mathcal{F}) + \mathbb{R}_t(i,\mathcal{G})\right) \end{pmatrix}\right\} \quad (22)$$

In this context, $\mathbb{Q}_t(i)$ denotes the Q-value associated with content '$i$' at the '$t^{th}$' epoch. $\mathbb{R}_t(i,_)$ signifies the corresponding reward gained from caching content '$i$'. The term "$[(x,y,z)_{A-UAV'\mathbb{X}'} = (x,y,z)_{MF-UAV}]$" defines the condition inside the *Iverson bracket*, taking the value 1 if micro-ferrying UAVs are within the communication range of A-UAV "$\mathbb{X}$" and 0 otherwise. The hyper-parameter "$\alpha_L$" governs the learning rate.

Initially, all the Q-values for contents start at zero, ensuring no prior information for the *Top-k* MAB agent and assigning equal importance to all contents for caching decisions. As the learning process advances, Q-values evolve, and the best contents, characterized by the highest Q-values, are cached. This approach aims to maximize the cumulative reward, subsequently enhancing the caching policy and, in turn, improving content availability.

The *Top-k* MAB agent faces a challenge due to the numerous content combinations $\binom{C}{k}$ it must sample for caching to get the best possible estimated values for all contents. An infrequent sampling results in weak reward distribution estimates, especially as the global content population ($C$) increases. To address this, $\epsilon$ and its decay rate are empirically chosen in the *$\epsilon$-Greedy* action selection policy [32]. To reduce policy dependence on $\epsilon$, an Upper Confidence Bound (UCB) strategy [31] is employed.

$$\mathbb{U}_t(i) = \mathbb{Q}_t(i) + \sqrt{\frac{\varsigma \log(t)}{\mathbb{m}_t(i)}} \quad (23)$$

The UCB, denoted as $\mathbb{U}_t(i)$, is calculated using the updated Q-value $\mathbb{Q}_t(i)$, controlling hyperparameter '$\varsigma$', and the number of times content '$i$' has been requested $\mathbb{m}_t(i)$. This strategy aids in content selection by favoring items with high reward potential and infrequent requests, thus promoting exploration without introducing an external exploration parameter $\epsilon$.

**Algorithm 2.** Caching policy at a A-UAV with *Top-k* MAB Learning

1. **Initialization:**
   a. $C$: Total contents in the system
   b. $C_A$: Caching capacity of an A-UAV
   c. $\mathbb{U}$: Size $|C_A|$ initialized with 0's (Q-table with UCB)
   d. $\alpha$: Learning rate for Q-table update
   e. $\varsigma$: Degree of exploration (in UCB)
2. **Load** A-UAV's cache with $C_A$ randomly chosen contents.
3. **while** True:
4. **Check** for learning epoch at A-UAV i.e., at $t^{th}$ epoch
5. **if** True **then do**
6. **for** $i = 0$ $to$ $length$(A-UAV cache size $C_A$) **do**
7. **Get** reward $\mathbb{R}_t(i, _)$ \\ according to Equations. 19-21
8. **Update** $\mathbb{U}_t(i)$ \\ from Equation. 23
9. **end for**
10. $value = \boldsymbol{copy}(\mathbb{U})$ \\ make a copy of UCB values

\\ Reload contents (Select arms)

11. **for** $i = 0$ $to$ $length$(A-UAV cache size $C_A$) **do**
12. $c_{max} = \boldsymbol{argmax}(value)$
13. **Load** $c_{max}$ to A-UAV
14. **Set** $value[c_{max}] = -\infty$
15. **end for**
16. **end if**
17. **end while**

The segment-based caching benchmarks in this study are used solely for theoretical comparisons and do not reflect how the *Top-k* MAB framework operates. These benchmarks represent deterministic edge cases for establishing performance bounds, whereas the learning-based *Top-k* MAB model is fundamentally more capable in handling fairness dynamically. Caching decisions are determined through a multi-dimensional reward structure that accounts for individual content popularity, inter-community content influence, and the global impact of caching choices. Unlike benchmarks that impose static constraints, the combinatorial nature of *Top-k* MAB inherently ensures fairness by dynamically adjusting caching priorities based on real-time content request patterns.

The Q-values computed in the combinatorial bandit model quantify content importance rather than predefined allocation constraints. Caching ability depends solely on UAV storage capacity that ensures fairness emerges as a function of storage limitations rather than arbitrary segment size constraints.

Furthermore, *Top-k* MAB's caching efficiency is entirely dependent on request generation patterns (i.e., sampling), unlike deterministic benchmarks that rely on predefined segment structures. Larger communities inherently generate higher request volumes which leads them to receive more cached content organically through bandit learning. This eliminates the need for manually imposed fairness constraints. Similar effects can be expected while content requests manifest urgency or emergency, for which the learning-based model adapts such that it can determine the importance of contents.

*Complexity Analysis*: To evaluate the computational feasibility of the *Top-k* Multi-Armed Bandit algorithm, we analyzed the complexity of its key components, including initialization, reward calculation, action selection, and the Q-value updates. These components collectively determine the scalability of the algorithm for large-scale UAV-assisted content dissemination systems.

The algorithm begins with an initialization phase where each A-UAV creates a Q-table to track the rewards associated with all $N$ contents. This initialization step, which is performed only once, has a time complexity of $O(N)$. Given the one-time nature of this step, its computational burden does not affect the scalability of the system in a major way. During each learning epoch, the algorithm computes rewards from $k$ arms selected from the total $N$ contents. This reward calculation, which occurs across $T$ epochs, involves evaluating the multi-dimensional reward structure. As $k \ll N$, the complexity of this step is $O(T.k)$, making it computationally light enough for real-time operations. The most computationally intensive step is the action selection process, where the Upper Confidence Bound (UCB) method ranks all $N$ contents to identify the top $k$ arms to cache. Sorting the contents at each epoch results in a complexity of $O(N \log N)$, which, over $T$ epochs, leads to a total complexity of $O(T.N \log N)$. While this step dominates the overall computational complexity, it is till bounded by logarithmic growth, thus keeping it scalable for large content pools. Finally, the Q-values of the selected arms are updated at each epoch based in the observed rewards. This update process has a complexity of $O(T.k)$, which remains manageable due to the small value of $k$ relative to $N$.

Combining these components, the overall time complexity of the *Top-k* MAB algorithm is $O(N) + O(T.k) + O(T.N \log N) + O(T.k)$. Simplifying this for $k \ll N$, the dominant term is $O(T.N \log N)$. Therefore, the overall time complexity is $O(T.N \log N)$, with the action selection step contributing the most significant computational cost. This complexity demonstrates the algorithm's suitability for large-scale systems, as its logarithmic scaling ensures efficiency even with large number of contents. Furthermore, the *selective caching* mechanism employed by MF-UAVs, which will be explained in a forthcoming subsection, with a linear complexity of $O\left(N_{Trajectory}.(C_{MF} + N_A)\right)$, complements the learning-based approach by ensuring effective content distribution without introducing significant computational overhead. Note that here $C_{MF}$ is the cache size of MF-UAVs, $N_A$ is the number of A-UAVs, and $N_{Trajectory}$ is the number of trajectory changes.

Thus, the overall complexity of the *Top-k* MAB algorithm is dominated by $O(T.N \log N)$, which makes it scalable for large content pools. The linear complexity of the MF-UAV caching mechanism further supports efficient operation in distributed settings. The proposed framework is computationally efficient, scalable and is well-suited for real-time UAV-assisted content dissemination in disaster-stricken environments.

There are a few limitations of the above learning based caching method. First, relying on MF-UAVs to ferry global content availability information makes learning slow. Especially so in large disaster areas with multiple communities. Second, communities with fewer users result in fewer content requests for the corresponding local A-UAVs. The problem is particularly compounded for the less popular contents for which the popularity reduces drastically following Zipf distribution (see Section 3.2). This lack of requests creates a learning challenge, leading to less accurate reward distribution estimates [33]. This results in unstable Q-values for less popular content. Finally, the *Top-k* MAB agent has to sample $\binom{C}{k}$ content combinations. This results in infrequent reward estimations with increasing $C$ (i.e., the total number of contents in pool), thus weakening the estimate of the reward distribution [33]. This leads to sensitive and unstable Q-values of contents.

These challenges can be mitigated by employing a Federated Multi-Armed Bandit (FedMAB) Learning approach. Such an approach involves integrating the *Top-k* MAB models from all A-UAVs. The mechanism is presented below.

### 6.2 Distributed Caching with Federated Multi-Armed Bandit

This mechanism applies the principles of Federated Learning [4, 29, 34, 35] to the UAV-caching scenario. Each A-UAV serves as a client [34] in Federated Learning, with its local model representing information about cached contents, cache hits [16], and content availability. Note that cache hits indicate how often a content cached in an A-UAV is requested and served within the user-species $TAD$, as defined in Section 3.2. The Q-table of the *Top-k* MAB agent serves as the model for each A-UAV.

MF-UAVs acts as model aggregators [34], chosen for their ability to access Q-tables of A-UAVs in their respective trajectories. They aggregate the acquired Q-tables aiming to improve the *Top-k* MAB model at each A-UAV. They receive the Q-tables from all A-UAVs in their trajectories, and send the aggregated model back to the A-UAVs. This aggregated model helps the A-UAVs to decide as to which content to cache based on the top '$|C_A|$' Q-values.

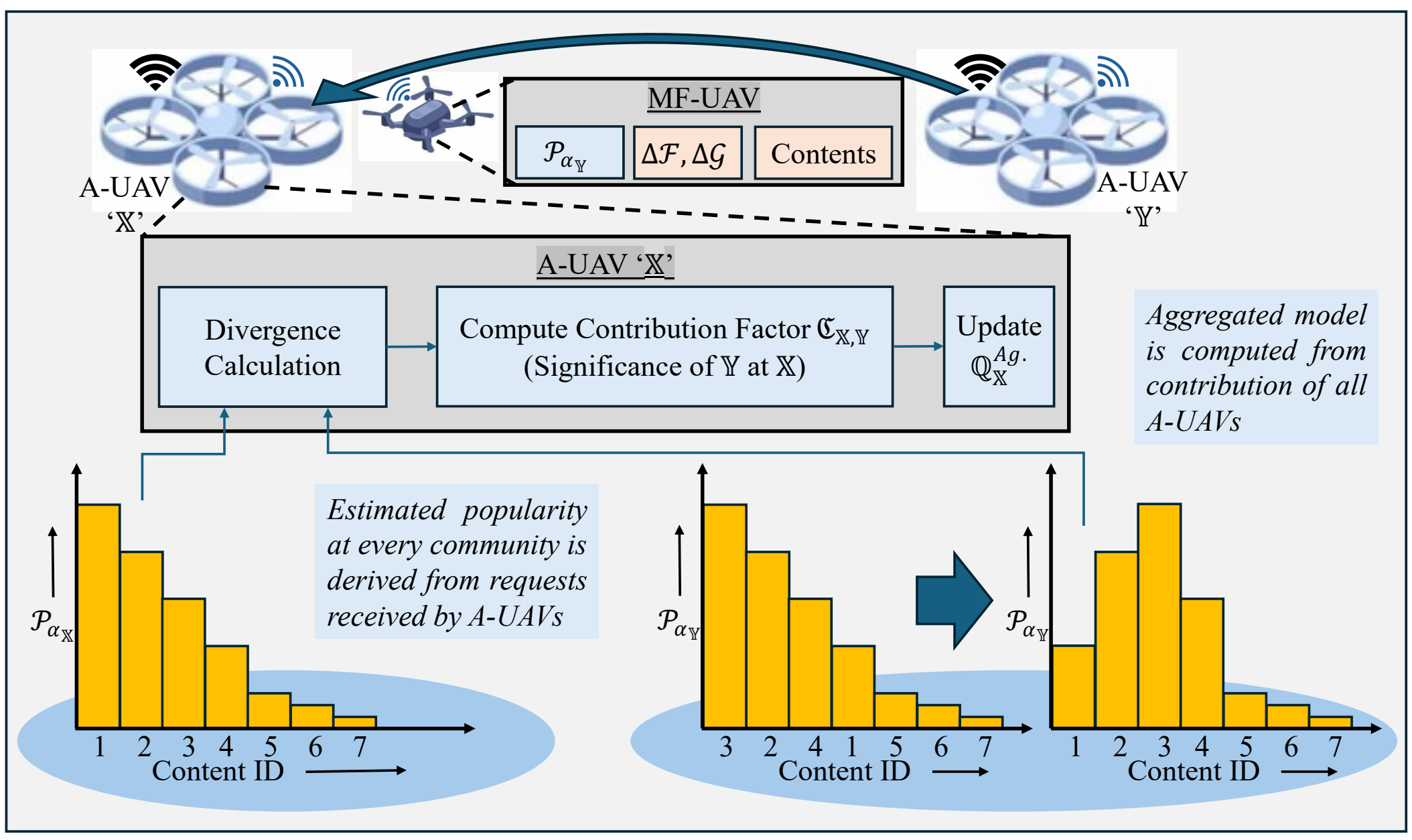

Fig. 5. Contribution factor for Federated Multi-Armed Bandit implementation at A-UAVs

As per the standard Federated Learning paradigm, Q-tables are initialized at the A-UAVs, and learning epochs are set based on the MF-UAVs' visiting frequencies. The learning epoch's dependance on MF-UAVs' visit is important to capture the rewards $\mathbb{R}(\mathcal{F})$ and $\mathbb{R}(\mathcal{G})$. Note that these rewards are associated to the ferried contents and their impact on global availability (refer Equations. 20 and 21). The Q-values of individual *Top-k* MAB agents are updated at each epoch, thus capturing the latest content request patterns and A-UAV caching decisions. This is termed as "*personal experience*" which is akin to the local training stage in Federated Learning. After gaining personal experience, an A-UAV's model is improved by exchanging information with its adjacent A-UAVs through the traveling MF-UAVs. To be noted that the quality of the aggregated model depends on the freshness of information ferried by the MF-UAVs. To be noted that cost constraints can reduce the deployment of high-cost UAVs, leading to less frequent information updates. Therefore, leveraging more number of affordable MF-UAVs within the UAV-assisted caching system ensures more consistent information collection, crucial for maintaining model accuracy under budgetary limitations.

Unlike weight matrix aggregation in neural networks in regression/classification models [35], Federated Multi-Armed Bandit involves the aggregation of Q-values. Each A-UAV's contribution during aggregation is determined based on its importance. This is synonymous to weights associated with devices in classical Federated Learning algorithm. Such contributions can be defined as *contribution factor* which is crucial during the aggregation process. It is calculated based on the similar between the estimated popularity distributions of contributing A-UAVs. The mechanism is shown using the following expressions:

$$\mathfrak{C}_{\mathbb{X},\mathbb{Y}} = [\rho - \mathfrak{D}_{KL}(\mathcal{P}_{\mathbb{X}}||\mathcal{P}_{\mathbb{Y}})] \Bigg/ \left[\sum_{\mathbb{l}=1}^{N_A} \left(\rho - \mathfrak{D}_{KL}(\mathcal{P}_{\mathbb{X}}||\mathcal{P}_{\mathbb{l}})\right)\right] \tag{24a}$$

$$\text{where}, \rho = \max\left(\sum_{\mathbb{l}=1}^{N_A} \mathfrak{D}_{KL}(\mathcal{P}_{\mathbb{X}}||\mathcal{P}_{\mathbb{l}})\right) \tag{24b}$$

Using the Equation. 25, the aggregated model can be shown as:

$$\mathbb{Q}_{\mathbb{X}}^{Ag.}(i) = \sum_{\mathbb{y}=1}^{N_A} \left(\mathfrak{C}_{\mathbb{X},\mathbb{y}} \times \mathbb{Q}_{\mathbb{y}}(i)\right) = \left[\sum_{\mathbb{y}=1}^{N_A} \left\{\left(\rho - \mathfrak{D}_{KL}(\mathcal{P}_{\mathbb{X}}||\mathcal{P}_{\mathbb{y}})\right) \times \mathbb{Q}_{\mathbb{y}}(i)\right\}\right] \Bigg/ \left[\sum_{\mathbb{l}=1}^{N_A} \left(\rho - \mathfrak{D}_{KL}(\mathcal{P}_{\mathbb{X}}||\mathcal{P}_{\mathbb{l}})\right)\right] \tag{25}$$

In Equation. 24, the contribution factor $\mathfrak{C}_{\mathbb{X},\mathbb{Y}}$ denotes the significance of A-UAV $\mathbb{Y}$'s model when the MF-UAV is at A-UAV '$\mathbb{X}$', where $N_A$ is the total number of A-UAVs. $\mathcal{P}_{\mathbb{X}}$ and $\mathcal{P}_{\mathbb{Y}}$ are content popularity distributions estimated at A-UAVs '$\mathbb{X}$' and '$\mathbb{Y}$', respectively. The KL divergence [37], denoted as '$\mathfrak{D}_{KL}(\mathcal{P}_{\mathbb{X}}||\mathcal{P}_{\mathbb{Y}})$', quantifies the distinction between these distributions. Thus, the term "$\rho - \mathfrak{D}_{KL}(\mathcal{P}_{\mathbb{X}}||\mathcal{P}_{\mathbb{Y}})$" represents how similar the content popularity distributions are near A-UAVs '$\mathbb{X}$' and '$\mathbb{Y}$'. To be noted that the term '$\rho$' is included in the expressions due to the unbounded nature of '$\mathfrak{D}_{KL}(\mathcal{P}_{\mathbb{X}}||\mathcal{P}_{\mathbb{Y}})$'.

Using the contribution factor from Equation. 24, the aggregated model is determined through Equation. 25. In this equation, $\mathbb{Q}_{\mathbb{X}}^{Ag.}(i)$ signifies the aggregated Q-value of content '$i$' at A-UAV '$\mathbb{X}$'. A higher importance is assigned to A-UAV "$\mathbb{y}$'s" model if its estimated content popularity distribution is more similar to that of A-UAV '$\mathbb{X}$', and vice versa.
Aggregating the Q-tables enhances the estimated reward associated with each content. In a learning epoch, the generated requests at a community might be insufficient for an accurate reward estimate at its A-UAV. Without model aggregation, A-UAVs end up with a weaker estimate of the reward distribution. Q-table aggregation, as proposed above, enhances the estimated rewards without requiring content request information from all the A-UAVs.

However, Q-table aggregation overlooks local popularity nuances when demand varies among the communities. This issue mirrors the personalization-generalization problem in Federated Learning [34, 36]. A-UAV's Q-table, updated with personal experiences using Equations. 22 and 23, is akin to a personalized (local) model, while the aggregated Q-table in Equation. 25 signifies a generalized (global) model. This paper employs weighted averaging to retain local popularity context while improving reward estimation. This can be expressed as:

$$\mathbb{Q}_{\mathbb{X}}^{Upd.}(i) = \omega_1 \mathbb{Q}_{\mathbb{X}}(i) + \omega_2 \mathbb{Q}_{\mathbb{X}}^{Ag.}(i) \tag{26}$$

In the given context, the weights $\omega_1$ and $\omega_2$ are critical in determining the influence of both local and global (aggregated) models in updating the model $\mathbb{Q}_{\mathbb{X}}^{Upd.}$. For the experiments in this paper, $\omega_1$ is empirically set to 0.99, indicating a strong preference for local content popularity, which is assumed to be relatively stable over time. However, a correct choice of $\omega_1$ is pivotal, especially in scenarios where the content popularity is dynamic. For such cases, the adaptive selection of $\omega_1$ is governed by:

$$\omega_1 = 1 - \left[\mathfrak{D}_{JS}\left(\mathcal{P}_{\mathbb{X}}^{t}, \mathcal{P}_{\mathbb{X}}^{t'}\right)/ln\,2\right];\ \omega_1\!: \mathfrak{D}_{JS}\left(\mathcal{P}_{\mathbb{X}}^{t}, \mathcal{P}_{\mathbb{X}}^{t'}\right) = \frac{1}{2}\left(\mathfrak{D}_{KL}(\mathcal{P}_{\mathbb{X}}^{t}||\mathcal{M}) + \mathfrak{D}_{KL}\left(\mathcal{P}_{\mathbb{X}}^{t'}||\mathcal{M}\right)\right) \tag{27}$$

Here, $\mathfrak{D}_{JS}\left(\mathcal{P}_{\mathbb{X}}^{t}, \mathcal{P}_{\mathbb{X}}^{t'}\right)$ is the Jensen-Shannon Divergence [37], indicating the dissimilarity between content popularity distributions at times $t$ and $t'$. $\mathcal{M}$ is the mean distribution calculated as $\mathcal{M} = \left(\mathcal{P}_{\mathbb{X}}^{t} + \mathcal{P}_{\mathbb{X}}^{t'}\right)/2$ and $\mathfrak{D}_{KL}(\mathcal{P}_{\mathbb{X}}^{-}||\mathcal{M})$ is the Kullback-Leibler divergence. A high $\omega_1$ value implies minimal change in content popularity over time.
The weight associated with the aggregated model $\omega_2$ in Equation. 26 can be expressed as:

$$\omega_2 = e^{-\beta_d t} \times \left(1 - \mathbb{Q}_{\mathbb{X}}(i)\right)/\beta_s \tag{28}$$

Here, $\beta_d$ and $\beta_s$ represent the weight decay factor and scaling factor, respectively. The parameters ensure that the contribution of global (aggregated) model reduces as learning progresses. The formulation of Equation. 28 reflects the decreasing relevance of the global model as learning advances, with an embedded regret component "$1 - \mathbb{Q}_{\mathbb{X}}(i)$". This idea is backed by the assumption that as learning progresses, the local models will reflect the true value of contents. Hence, the expressions for weights $\omega_1$ and $\omega_2$ from Equations. 27 and 28 can be replaced in Equation. 26:

$$\mathbb{Q}_{\mathbb{X}}^{Upd.}(i) = \left[1 - \left\{\mathfrak{D}_{JS}\left(\mathcal{P}_{\mathbb{X}}^{t}, \mathcal{P}_{\mathbb{X}}^{t'}\right)/ln\,2\right\}\right] \times \mathbb{Q}_{\mathbb{X}}(i) + \left[\frac{e^{-\beta_d t}}{\beta_s} \times \left(1 - \mathbb{Q}_{\mathbb{X}}(i)\right)\right] \times \mathbb{Q}_{\mathbb{X}}^{Ag.}(i) \tag{29}$$

The hyper-parameters '$\beta_d$' and '$\beta_s$' should be empirically optimized to maintain local relevance in content popularity, particularly in heterogeneous environments.

This framework integrates Federated Learning with *Top-k* Multi-Armed Bandit principles for high-performance caching in autonomous UAVs (A-UAVs). It leverages model aggregation at MF-UAVs and updates A-UAV models with a balance of personalized and aggregated data. The updated Q-table is used for caching decisions, thus prioritizing contents with the highest Q-values for enhanced content availability across all user communities.

The dependency on MF-UAVs for model aggregation does not introduce significant bottlenecks, as multiple MF-UAVs operate in parallel to facilitate decentralized updates across different A-UAVs. This naturally distributes the computational load by preventing a single UAV from becoming an aggregation bottleneck. Additionally, the contribution-based weighting mechanism (Equation 24) ensures that only relevant updates are merged which reduces unnecessary model exchanges and further optimizes aggregation efficiency.

For even greater scalability, an alternative approach could involve hierarchical aggregation, where A-UAVs first perform localized model updates before forwarding refined models to MF-UAVs for global merging. This multi-level aggregation structure could reduce communication overhead in networks with a larger number of UAVs that can further enhance system scalability. While the current approach remains effective under the evaluated network conditions, future work can explore hierarchical aggregation strategies to optimize performance in extremely large deployments.

The proposed FedMAB framework ensures privacy in federated learning-based caching by limiting shared information to content popularity distribution parameters and Q-values, which serve as reward distribution estimates. Unlike traditional federated learning approaches that involve direct transmission of graph-based models which represents user data or raw request logs, FedMAB does not share individual user content requests, behavioral patterns, or identifiable data. The aggregation process exchanges only learned model parameters relevant to content caching decisions which makes the framework inherently privacy-preserving. Since Q-values represent estimated rewards from caching actions rather than explicit user information, the system effectively optimizes caching policies without exposing sensitive data. Furthermore, these Q-values undergo aggregation at MF-UAVs, where updates are computed based on local learning experiences, without requiring access to underlying user-generated request data.

This decentralized learning mechanism ensures that UAVs collaboratively refine caching strategies while preserving privacy. Since no direct content request logs or personal information are transmitted, this structure ensures secure model sharing which minimizes the risk of data leakage while maintaining collaborative learning efficiency.

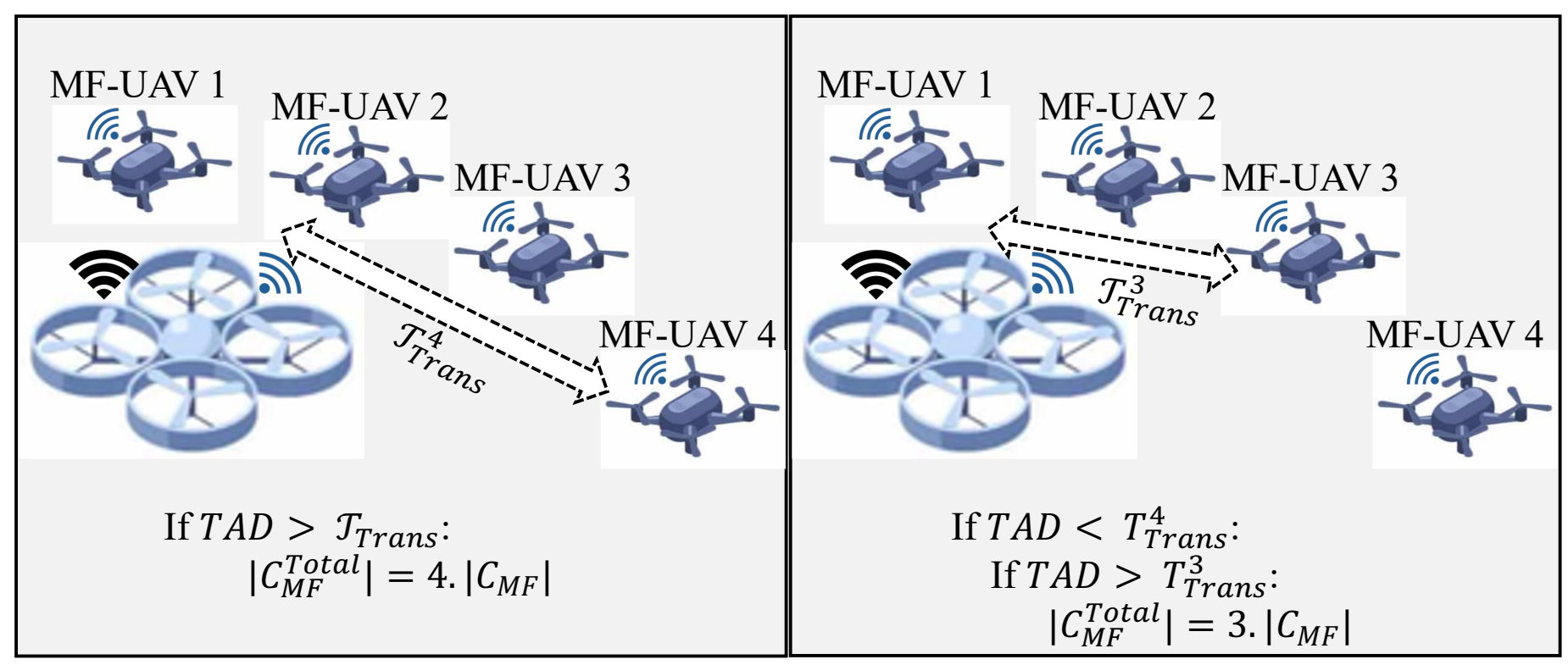

Fig. 6. Increase in collective caching capacity of MF-UAVs through Selective Caching

### 6.3 Selective Caching at Micro-Ferrying UAVs (MF-UAVs)

The role of MF-UAVs is to ferry contents from the previously visited A-UAVs to the future visiting A-UAVs such that the future visiting A-UAVs get the benefit of contents cached at other A-UAVs. Ideally, the purpose of the MF-UAVs is to ferry around a subset of $C_E^{total} + N_A.(1-\lambda).C_A$ number of contents stored across $N_A$ number of A-UAVs (see Section 5.1). However, such implementation leads to replication of all ferried contents, resulting in underutilized cache space at the MF-UAVs. Due to the limitation of per-MF-UAV caching space (i.e., $C_{MF}$), their caching policy should be jointly determined based on their trajectories, learnt caching policy at the A-UAVs, content request patterns, and the tolerable access delays ($TADs$) associated with the contents to be cached. A "*Selective Caching*" mechanism as the MF-UAV caching policy is explained in the pseudocode below.

---

**Algorithm 3.** Selective Caching Algorithm MF-UAV with

---

FedMAB caching at A-UAV

1. **Input:** Total A-UAVs in its trajectory, $TAD$, next A-UAV '$\mathbb{X}$', present A-UAV '$\mathbb{X}-1$'
2. **Output:** $C_{MF}$ contents for MF-UAV '$\mathcal{Y}$'
3. Caching at A-UAVs using FedMAB policy // Equations. 19-29
4. **while** True:
5.     **if** MF-UAV leaving for next A-UAV '$\mathbb{X}$' **then do**
        // Contents that are not in the future visiting A-UAV
6.         **Update** ferrying content knowledge
        // Function call from the present A-UAV '$\mathbb{X}-1$'
7.         **Call** content-wise_TAD ( )

// Present A-UAV sends MF-UAV visiting frequency
8. **Call** MF-UAV_visiting_frequency ( )
// Check what content the last MF-UAV ferried
9. **Call** Check_previous_MF-UAV_roster ( ) **Return** roster contents with respective TADs
// Compute request interval for last MF-UAV roster
10. **Calculate** least popular content's request interval
11. **Check** if request time is less than its TAD and MF-UAV visiting duration
12. **if** True **then do**
13. Cache same roster
14. **else**
15. Cache next best roster
16. **end if**
17. **Check** if other MF-UAVs flying with MF-UAV '$\mathcal{Y}$'
18. **for** $l$ = 0 to $length$(MF-UAVs flying together) **do**
19. **for** $k = 0$ $to$ $length$(A-UAV '$\mathbb{X}$' cache $C_A^{\mathbb{X}}$) **do**
20. **Check** if $k$ in $C_{MF}$ cache space of MF-UAV '$\mathcal{Y}$'
21. **if** True **then do**
22. Replace '$k$' with highest value content from $C_A^{\mathbb{X}-1}$ not cached in MF-UAV '$\mathcal{Y}$' and A-UAV '$\mathbb{X}$'
23. **end if**
24. **end for**
25. Cache next best roster
26. **end for**
27. **end if**
28. **Update** next A-UAV '$\mathbb{X}$', present A-UAV '$\mathbb{X}-1$'
29. **end while**

In Algorithm 3, the process of selective caching is described in detail. Consider a situation in which an MF-UAV '$\mathcal{Y}$' is ready to leave the A-UAV '$\mathbb{X}-1$'. Before caching contents, it needs the following information from A-UAV '$\mathbb{X}-1$'; 1) what are the contents eligible for ferrying to A-UAV '$\mathbb{X}$'; 2) what is the MF-UAVs visiting frequency; 3) what roster of ferrying content did the last MF-UAV ferry, where roster is the grouping of contents based on their popularity or value; 4) are the next roster contents likely to be requested within the given $TAD$; and 5) are MF-UAVs flying in groups. Based on these information, MF-UAV '$\mathcal{Y}$' selectively caches contents which helps in maintaining diversity in the contents cached by all MF-UAVs in its vicinity. This means, if MF-UAVs are flying in groups or traversing in close proximity from each other, they ferry contents from consecutive rosters. To be noted that the size of a roster is same as an MF-UAV's cache size. Therefore, if subsets of MF-UAVs are considered collectively as a group of $N_{MF}^{G}$ (group size), then the number of contents cached by the group is $N_{MF}^{G} \times C_{MF}$. Such selective caching policy at MF-UAVs ensures content availability maximization by avoiding redundant content replication.

### 6.4 Enhancing Federated Learning with Controlled Latency

The use of A-UAVs equipped with federated multi-armed bandit (FedMAB) learning algorithms offers a promising avenue for adaptive learning and decision-making based on user demands and network conditions. However, the model aggregation nature of FedMAB, while enhancing content delivery services, can inadvertently diminish the benefits of selective caching strategies. Especially so when such a strategy is crucial for managing a UAV-network's storage resources effectively.

To address this, a nuanced latency approach that integrates Federated Multi-armed Bandit learning at A-UAVs with selective caching at MF-UAVs is proposed. This approach maintains the integrity and benefits of both federated learning at A-UAVs and selective caching at MF-UAVs by introducing controlled latency into the A-UAVs' learning cycles.

*Mechanism Details:* The modified FedMAB learning algorithm with latency introduces a deliberate delay in the divergence-based weighted computation updates of A-UAVs. In simpler terms it adds a delay between the *Top-k* MAB update and the model aggregation at A-UAVs. This delay is managed through a *latency_counter*, which tracks the number of learning epochs elapsed since the last federated learning update. Only when this counter exceeds a predefined threshold, $T_L$, does the A-UAV proceed with its learning and cache update process via federated learning (refer to Equations. 24-29). This controlled latency allows MF-UAVs more time for data analysis and informed decision-making regarding selective caching.

During the latency period, A-UAVs continue to collect data, learn via *Top-k* MAB agents, and perform their regular operational functions. However, they postpone the federated learning cycle's execution, allowing MF-UAVs to assess and analyze the cached content across various A-UAVs. MF-UAVs can then identify which contents are likely to be in higher demand and ensure their availability by ferrying them between A-UAVs. This synchronization of learning with the mobility patterns of MF-UAVs enables more strategic and informed decisions regarding content caching and distribution.

**Algorithm 4.** Federated Multi-Armed Bandit Learning with Strategic Latency for A-UAVs

1. **Input**:
    a. $C$: Total contents in the system.
    b. $C_A$: Caching capacity of an A-UAV.
    c. $T_L$: Latency Threshold.
2. **Initialization**:
    a. **Set** *latency_counter* to $0$ for each A-UAV
    b. **Initialize** each A-UAV's cache with randomly selected $C_A$ contents
    c. **Set** Q-values for all content to 0.
// These values help track content demand.
    d. **Define** learning rate ($\alpha$) and exploration parameter ($\varsigma$)
3. **Main Loop**:
4. **While** the system is running:
5. **Check** if it's time (current epoch) for a learning update
// This could be determined by MF-UAV flight time
6. **Calculate** reward $\mathbb{R}_t(i, _)$ for content $i$ in A-UAV
7. **Update** the Q-value for all cached contents using MAB
// Based on calculated reward and the learning rate ($\alpha$)
8. **If** *latency_counter* >= $T_L$ **then**:

9. **Compute** Divergence-based Weights
10. **Update** Q-values using Equations 24-29
11. **Reset** *latency_counter* to 0
    // Indicating an update has been completed.
12. **Else If** *latency_counter* <= $T_L$ **then**:
13. **Increment** *latency_counter* by 1
    // This delays the Federated learning update cycle
14. **Copy** the Q-values to a temporary list for manipulation
15. **For** each slot in the A-UAV's cache:
16. **Select** not cached content with the highest Q-value
17. **Update** the cache to include this content
    // Replace the least demanded content if necessary.
18. **Update** the selected content's Q-value to $-\infty$
    // In the temporary list to avoid reselection
19. **Repeat** steps 4-18 for an adaptive system

This latency-based approach enhances content availability across the network and optimizes the use of network resources, ensuring a balance between learning efficacy and caching efficiency. By integrating the dynamic learning capabilities of A-UAVs with the selective caching strategies of MF-UAVs, the system becomes more resilient, efficient, and user-centric.

Table 1: Default Values for Model Parameters

| # | Variables | Default Value |
|---|---|---|
| 1 | Total number of contents, $C$ | 2000 |
| 2 | Number of A-UAVs, $N_A$ | 4 |
| 3 | Number of MF-UAVs, $N_{MF}$ | 8 |
| 4 | A-UAV's Cache space (content count), $C_A$ | 200 |
| 5 | MF-UAV's Cache space, $C_{MF}$ | 25 |
| 6 | Poisson request rate parameter, $\mu$ (request/sec) | 1 |
| 7 | Hover rate of MF-UAV, $R_{Hover} = T_{Hover}/T_{Trajectory}$ | $1/6$ |
| 8 | Transit rate of MF-UAV, $R_{Transit} = T_{Transit}/T_{Trajectory}$ | $1/12$ |
| 9 | Zipf parameter (Popularity), $\alpha$ | 0.4 |
| 10 | Micro Ferrying UAV Trajectory | Round-robin |

## 7 EXPERIMENTAL RESULTS AND CONTENT DISSEMINATION PERFORMANCE

Simulation experiments were conducted to evaluate the performance of the proposed FedMAB learning-based caching mechanism and selective caching at micro-ferrying UAVs. An event-driven simulator was used to generate content requests, maintaining intervals between events according to an exponential distribution and following a Zipf popularity distribution (see Equation. 1). To account for variations in content popularity across different communities, contents were swapped with a predetermined probability [26], and differences between sequences were maintained using the Smith-Waterman Distance [38]. The default system parameters for the FedMAB-based caching and cache pre-loading policies are provided in Table I.

In the simulation, the impact of lateral link range on content dissemination has been implemented. An MF-UAV begins serving content upon entering the WiFi transmission range of a community, even before reaching its boundaries. The duration during which the MF-UAV starts transmitting content, denoted as $\Delta t_{comm}$, is influenced by its transit speed. If $\Delta t_{comm}$ is significantly shorter than the Poisson-distributed content request generation time ($T_{req}$), the adjusted hover time remains approximately the same ($\hat{T}_{Hover} \approx T_{Hover}$). Conversely, if $\Delta t_{comm}$ is comparable to or exceeds $T_{req}$, the adjusted hover time increases to $\hat{T}_{Hover} \approx T_{Hover} + \Delta t_{comm}$, while the transit time decreases to $\hat{T}_{Transit} \approx T_{Transit} - \Delta t_{comm}$.

The performance of the proposed mechanism was evaluated using the following metrics:

*Content Availability* ($P_{avail}$): This is the ratio of cache hits to generated requests within a tolerable access delay. Cache hits refer to content provided to users from the UAV-cached content without needing a download. Content availability indirectly reflects the content download cost of the system.

*Cache Distribution Optimality (CDO):* This metric assesses the optimality of the learned caching policy in terms of the caching sequence. The Jaro-Winkler Similarity (JWS) [39] measures CDO by comparing the similarity between the content sequence from the learned caching policy and the cache pre-loading sequence. It considers the number of matches, required transpositions, and prefix similarity of both sequences. A normalized similarity measure, where 1 indicates optimal caching and 0 indicates non-optimal caching, is used.

*Access Delay* ($AD$): Performance of FedMAB model and selective caching policy for micro-ferrying UAVs is also evaluated based on the access delay which is the end-to-end delay between the generation of content request and its provisioning from the cached contents in the UAVs. This paper reports the epoch-wise average access delay to show the improvement in caching policy as learning progresses.

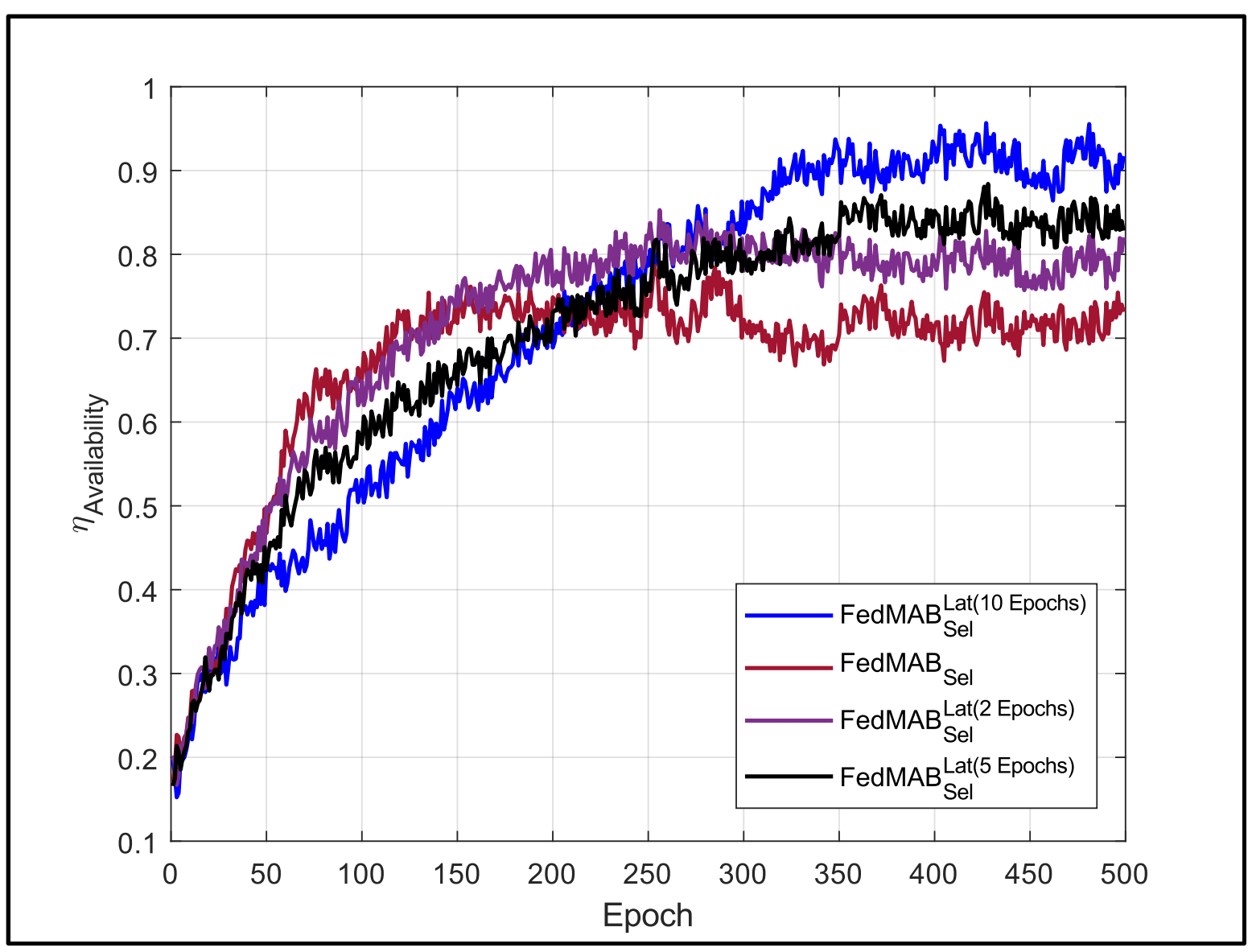

Fig. 7. Increase in content availability by controlling the learning latency in Federated Learning aided caching policy.

### 7.1 Effect of Controlled Latency Induced Federated Learning on Content Availability

To understand the applicability of the proposed FedMAB-based caching policy along with selective caching, experiments were conducted with different durations of controlled latency. This is achieved with caching policies learnt through models that update with different levels of latency. Each MAB model uses a hybrid exploration strategy including both UCB and $\epsilon$-greedy, where the degree of exploration is set to $\alpha_u = 2$. Also, to show the effectiveness of selective caching at micro-ferrying UAVs (MF-UAVs), *TAD* Ratio $R_{TAD}$ for contents $\{51 - 75\}$ are kept lower than the default $R_{TAD}$ i.e., $1/8$ . To be noted that *TAD*s are represented as a ratio with respect to trajectory time ($T_{Trajectory}$) to ensure generalizability of the proposed algorithms. Fig. 7 shows the convergence behavior of the learnt caching policy with FedMAB model at the A-UAVs, and selective caching at the MF-UAVs. The comparison emphasizes on the effects of controlled latency.

The convergence behavior is shown in terms of relative content availability, which is the ratio between content availability achieved using the proposed method and the deterministic baseline method from Equation 13-17. The key outcomes are given below. First, the best content availability achieved is with the maximum induced latency while implementing FedMAB to learn caching policy. This parameter controls the application of divergence-based weight computation, eventually the aggregation of the *Top-k* MAB (refer Equations 24-29 and Algorithm 4). Second, the promptness in learning behavior is more apparent in the models with least latency or no latency. However, the converged learning performance is subpar, and it can be seen via the attained content availability. Third, the learning progression is inversely proportional to the controlled latency for aggregation, whereas the learning performance is directly proportional to it. For least controlled latency, the individual model's epoch-wise reward estimate $\mathbb{E}[\mathbb{R}] \neq \mathbb{r}^*$ is weak due to limited content requests experienced within an epoch's duration. Here, $\mathbb{R}$ is the reward received during $t^{th}$ epoch and $\mathbb{r}^*$ is the true reward. Also, due to the mobility of the MF-UAVs, the accessibility of ferry and global content availability information can't be guaranteed, leading to a weak and sensitive estimated reward. On the contrary, with a high controlled latency, the individual model's reward is substantially stable i.e., $\mathbb{E}[\mathbb{R}] \approx \mathbb{r}^*$. Additionally, due to the induced latency for model aggregation, the content availability information from adjacent communities can be accessed via MF-UAVs with high likelihood. This leads to a better overall reward estimate, therefore improving content caching policy. However, the explicit introduction of latency to the learning algorithm makes the model update process sluggish, which can be seen in Fig. 7.

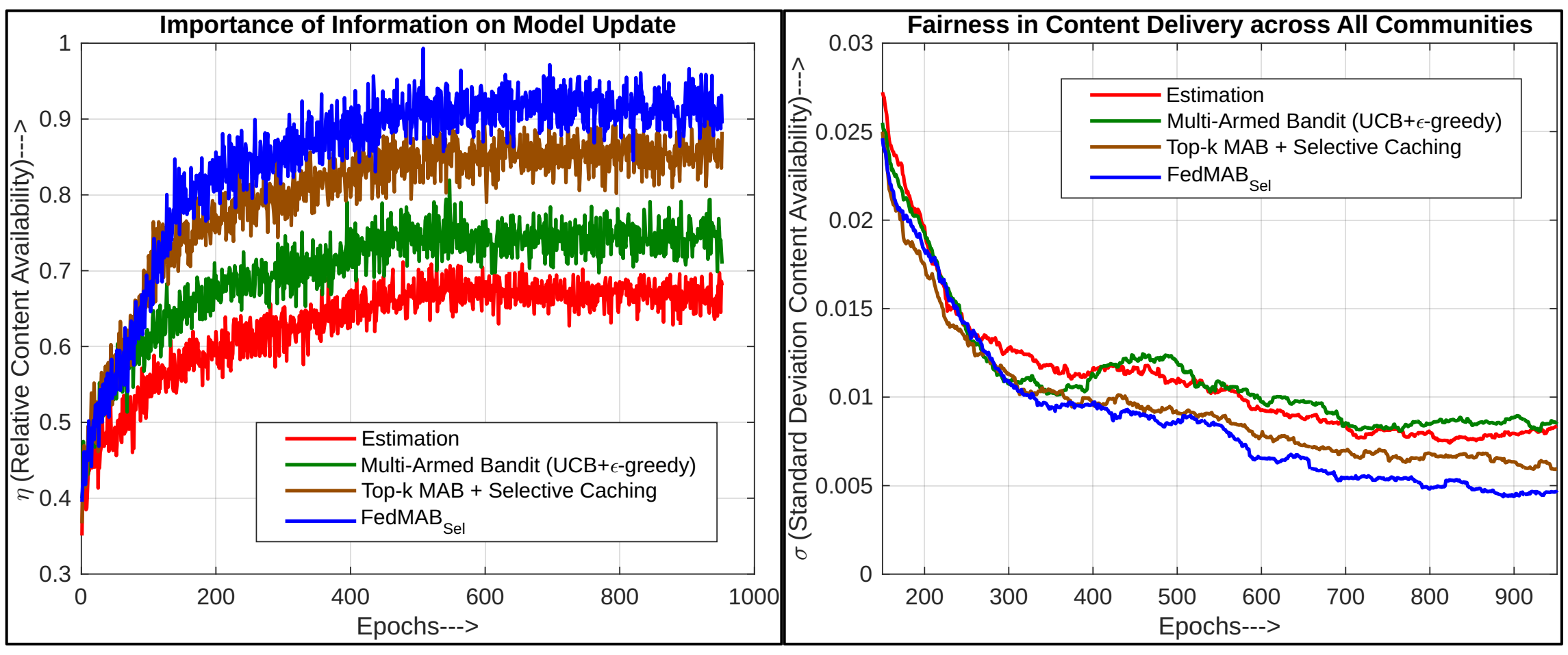

Fig. 8. (Left) Evolution of learning-based caching policy with information sharing; (Right) Uniformity of performance at all A-UAVs.

### 7.2 Evolution of Learning based Caching Policies and their Impacts

The evolution of the learning-based caching policies designed in this paper and their comparison are shown in terms of relative content availability. The observations from Fig. 8 are as follows. First, the figure shows that by employing FedMAB model along with selective caching, a caching policy can be learnt which can provide content dissemination performance closer to the benchmark performance [26]. The benchmark performance, using Value-based Caching, is calculated with the aid of *apriori* information on content popularity and takes into consideration the heterogeneity in user demand (see Equations 9-12 ). The proposed FedMAB algorithm is able to leverage the multi-dimensional reward structure and divergence-based weighted aggregation to account for heterogeneity [40, 41], as explained in Equations. 19-29, to learn the caching policy on-the-fly (see Section 6.1 and 6.2). Second, the selective caching policy at micro-ferrying UAVs leverages the shared information between themselves and with the A-UAVs to boost the content availability closer to the benchmark performance by approximately 20% (see Fig. 8). It utilizes the currently visiting A-UAV's caching information and the preceding MF-UAV's caching decision to algorithmically select its own contents for caching, which is also shown in Fig. 6. Such selective caching will reduce the redundancy of multiple copies of the same content available through multiple sources at the same time. Third, the difference in the efficacy and limitations of selective caching can be observed in Fig. 7 and 8, where caching decisions at MF-UAVs differ due to the model aggregation in both scenarios. The effectiveness of controlled latency can be seen here in Fig. 8, where the benefits of divergence-based weighted aggregation is preserved along with leveraging the pros of selective caching. Fourth, when the agent uses UCB exploration strategy, during the initial learning epochs the content availability increases promptly due to high upper confidence value of all contents, which avoids excessive exploitation. This is due to low sampling of requests. As learning progresses, the sparse request for unpopular contents keeps the upper confidence value high which maintains consistent exploratory behavior. Fig. 8 shows that such exploration strategy alone helps to boost the content availability closer to the benchmark performance by approximately 10% more than popular estimation-based methods [22]-[25]. Finally, the standard deviation across the performances of all A-UAVs is recorded, which shows the progression of the learning-based caching policy. Note that $\text{FedMAB}_{\text{Sel}}$ shows lowest standard deviation, which shows highest level of fairness in the performance. Here, the $\sigma$ is computed as average of 150 learning epochs. Also, contrary to the performance behavior of the MAB algorithm with hybrid action selection strategy, it shows more nonuniform increase in performance with respect to estimation-based methods. This can be attributed to intermittent accessibility of MF-UAVs, therefore limiting information access. This behavior is not seen in both learning variants with selective caching as the caching information is spanned across multiple MF-UAVs.

*Discussion*: $\text{FedMAB}_{\text{Sel}}$ allows maximum evolution of the caching policy such that increased $TAD$ is leveraged to achieve highest content availability. The performances of *Top-k* MAB with Selective caching and multi-dimensional reward structure follows in that order. These observations can be used to deepen the understanding of the components of $\text{FedMAB}_{\text{Sel}}$. With high $TAD$, the ferrying and global reward i.e., $\mathbb{R}(i,\mathcal{F})$, and $\mathbb{R}(i,\mathcal{G})$ respectively, brings the estimated reward $\mathbb{R}_T$ closer to the true mean, as it allows more time for the MF-UAVs to transit before the request expires. Furthermore, the effectiveness of the selective caching algorithm boosts with high $TAD$, since it allows more MF-UAVs to collaborate allowing them to avoid caching copies of same contents amongst themselves. The last component of $\text{FedMAB}_{\text{Sel}}$, that is the divergence-based weighted updates of the models allows each A-UAV to have explicit knowledge of the content popularity at adjacent communities, therefore avoiding content replication at A-UAVs. The consolidation of these components of $\text{FedMAB}_{\text{Sel}}$ results in increased content availability with the primary objective of this work. Note that a high value of $TAD$ also allows for unconstrained application of controlled latency, which proves to be beneficial in boosting content availability as shown in Fig. 7 and 8.

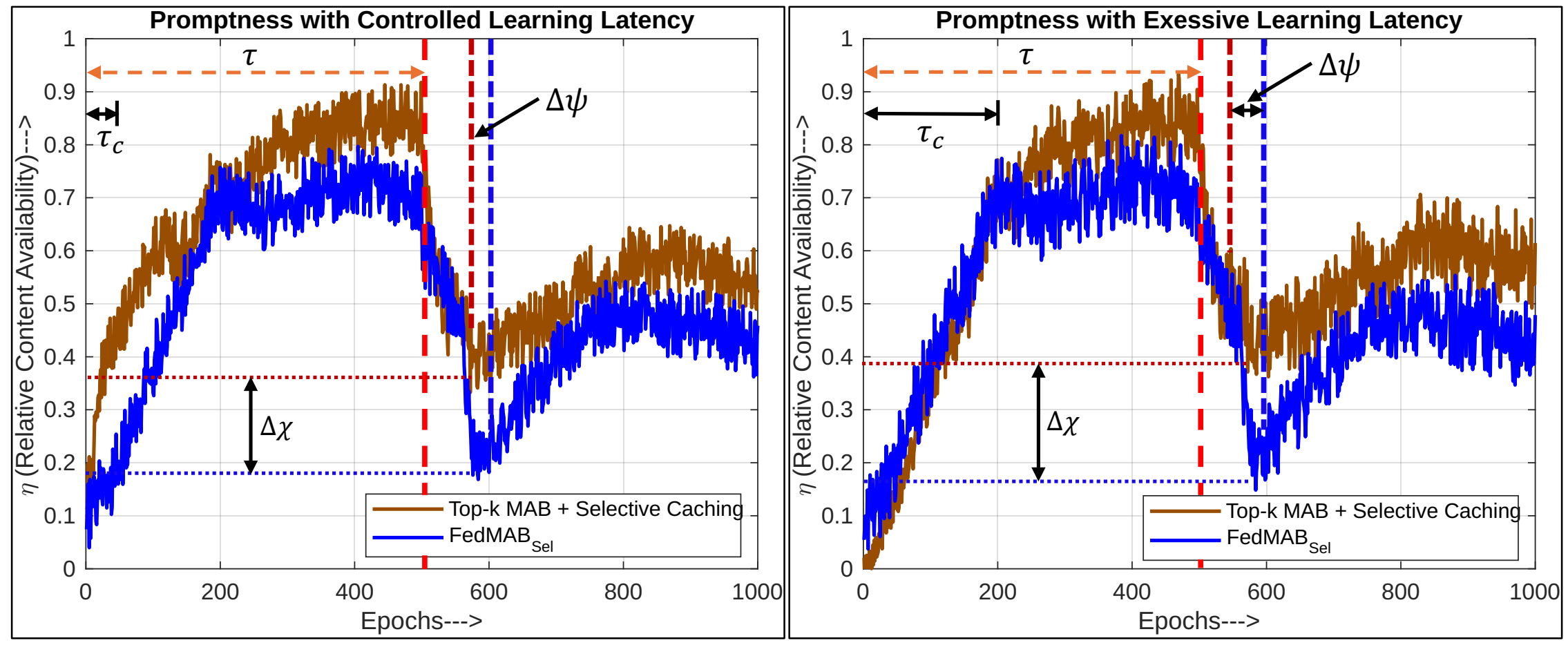

Fig. 9. Balance between reactiveness and performance of $FedMAB_{Sel}$ caching policy in case of time-varying user preferences

### 7.3 Adaptability with Changing User Preferences

The adaptability of the proposed learning-based caching mechanisms is further emphasized in Fig. 9. It showcases that the ability of $\mathrm{FedMAB_{Sel}}$ approach to learn the caching policy in a setting where the user preference changes over time. It goes on to highlight the reactive nature of the $\mathrm{FedMAB_{Sel}}$, where content availability increases more promptly compared to the standalone *Top-k* MAB implementation or any of its predecessors. Note that dynamic user preference patterns are simulated using Smith-Waterman Distance-based sequence swapping [38] and changing the Zipf parameter (refer Equation 1). Moreover, the comparison between the reactiveness of the learning-based caching polices are depicted in terms of 3 different measures, namely reactiveness time ($\psi$), lowest performance point ($\chi$) and crossover ratio ($\zeta$). Reactiveness time ($\psi$) captures the time taken for the system to start improving its performance after the demand scenario change. $\chi$ represents the lowest point in performance after the demand scenario change, just before the system begins to recover. Crossover ratio ($\zeta$) represents the ratio of the time before which one algorithm's performance surpasses another (i.e., $\tau - \tau_c$), relative to the time constant ($\tau$ which is the duration of the fixed demand scenario). Here, $\tau_c$ refers to the time when an algorithm's performance surpasses another. Therefore, crossover ratio $\zeta$ can be expressed as $\zeta = \frac{\tau - \tau_c}{\tau}$. For interpretability, the case where performance of an algorithm doesn't surpass its predecessor, $\tau_c = \tau$. This indicates that there is no relative improvement in performance within the time constant $\tau$. The performance seen with $\mathrm{FedMAB_{Sel}}$ exhibit relatively lower values for $\psi$ and higher values for $\chi$ with any level of controlled latency in $\mathrm{FedMAB_{Sel}}$. This indicates the promptness of the proposed caching method as compared to its predecessors. Crossover ratio $\zeta$, on the other hand, shows a more nuanced observation. For controlled latency of 2 epochs, $\zeta$ is high but it reduces for latency of 10 epochs, although with improved relative performance. This shows that a high controlled latency in divergence-based weighted updates for $\mathrm{FedMAB_{Sel}}$ can improve performance significantly, but it comes with a cost of the model's reactiveness. Therefore, a realistic assumption on the dynamic nature of the content demand pattern suggests that for user preferences with high time constant $\tau$, the reactiveness of the $\mathrm{FedMAB_{Sel}}$ is relatively high as compared to the learning-based caching mechanisms discussed above.

Note that while the Zipf model is used to represent content request generation patterns, the proposed Federated Multi-Armed Bandit (FedMAB) framework remains agnostic to the specific distribution governing user requests. The caching decisions are purely data-driven that relies on observed request patterns rather than prior assumptions about the underlying distribution. If content requests were generated according to an alternative distribution, such as Normal, T, or Beta, the learning process would naturally adjust caching policies to match the observed demand structure. The adaptability of FedMAB ensures that the system remains effective under diverse content popularity dynamics, therefore optimizing caching decisions based on real-world user behavior rather than a predefined statistical model.

*Discussion*: The choice of *TAD* for the experiments is such that it is less than the hovering and transiting duration together. This is done to emphasize the reactive nature of the algorithm by constraining the allowed duration for a request, before it is served via download. It should be highlighted that keeping the *TAD* too high allows the MF-UAVs to reduce the caching frequency of those contents. On the contrary, for very low *TAD*, the model overestimates the value of those contents leading to them being cached at A-UAVs allowing ready availability.

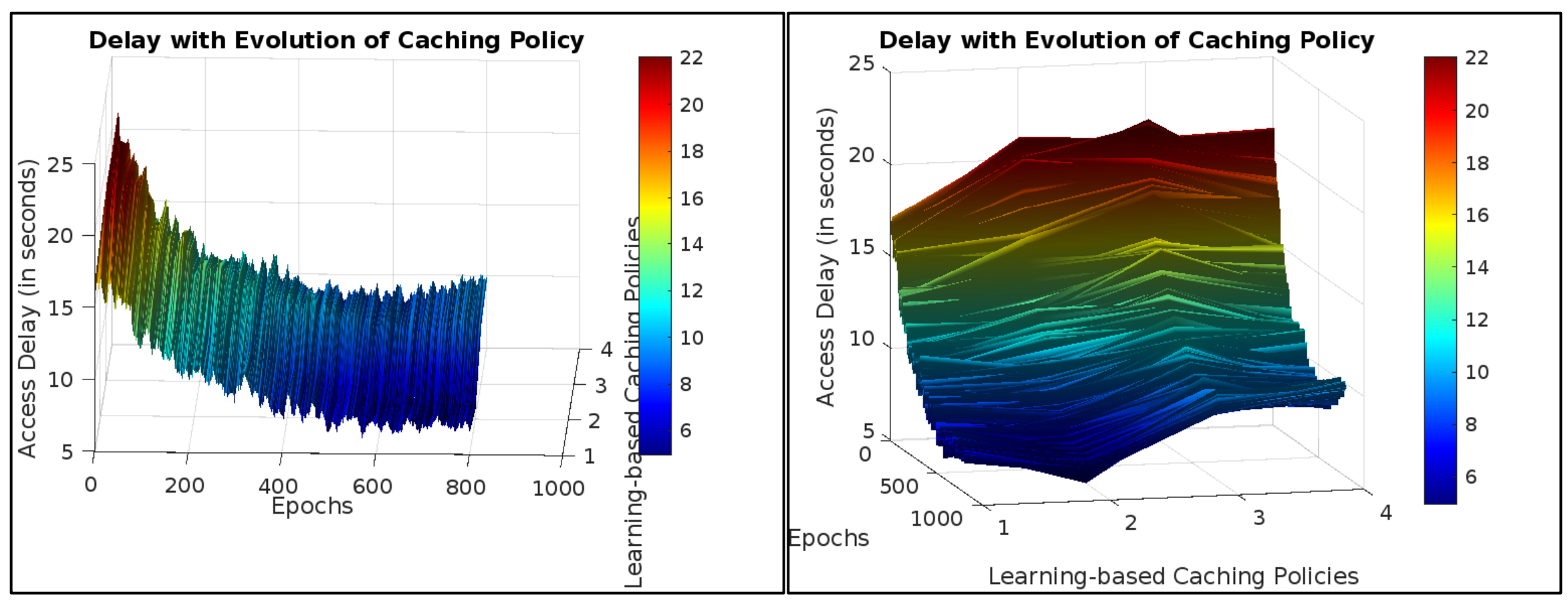

Fig. 10. Access delay as a determinant for the choice of learning-based caching policy (Two viewing perspective)

### 7.4 The Interplay Between Learning Latency and Content Access Delay

The choice of learning-based caching policy with respect to the access delay has been highlighted in Fig. 10. This figure emphasizes on the various components, namely multidimensional reward structure, selective caching and divergence-based weighted aggregation, the amalgamation of which leads to the proposed $\text{FedMAB}_{\text{Sel}}$ caching policy. Additionally, it also scrutinizes the behavior of these components under the influence of shared information in varying learning-based caching scenarios. The observations are as follows. First, both Fig. 8 and 10 shows that with increase in shared information content availability and access delay improves irrespective of the caching policy used. However, the efficacy of different versions of the learning-based caching policies varies. Second, Fig. 10 demonstrates the effect on access delay while applying different versions of the learning-based caching policy as learning progresses. It can be seen that for each learning-based caching policy the delay decreases with increase in epochs, which is intuitive. Along with providing more contents from the hierarchical UAV-aided content dissemination system, the delay decreases since more relevant contents are stored at A-UAVs. Third, with the implementation of multi-dimensional reward structure and the hybrid action selection strategy, access delay decreases, since the relevance of the contents cached in both A-UAVs and MF-UAVs improve. Finally, it can be that as the learning-based caching policy evolves the access delay reduces till a certain point. However, an increase in delay can be seen for $\text{FedMAB}_{\text{Sel}}$ based policy. The reasons are multifaceted. When the model evolves from standalone MAB to *Top-k* MAB with multi-dimensional reward structure, the caching policy of the A-UAVs improve leading to high value content being cached at A-UAVs. This results in better content being ferried via the MF-UAVs without adding significant delay. $\text{FedMAB}_{\text{Sel}}$ along with improving the caching policy for A-UAVs improves the policies for MF-UAVs jointly, which allows more content to be ferried from adjacent communities before exhausting the requests lifetime. This increases the dependance on the hierarchical UAV-aided dissemination system for content provisioning. Therefore, an increase in access delay is observed along with a boost in content availability, which is the primary objective of this work (refer Section 4).

*Discussion*: The proposed framework inherently mitigates latency through a combination of controlled learning updates, demand-aware caching policies, and dynamic UAV coordination. As detailed in Algorithm 4, the system regulates the timing of federated model updates to balance responsiveness and caching stability which ensures that UAVs do not prematurely overwrite cached content while maintaining adaptability to real-time requests. The incorporation of Tolerable Access Delay (*TAD*) constraints ensures that content with lower latency requirements is prioritized at A-UAVs, while content with more relaxed delay tolerance is efficiently transferred via MF-UAVs.

Additionally, high-demand conditions naturally strengthen the learning process of the Federated Multi-Armed Bandit (FedMAB) framework. Increased accessibility of MF-UAVs in such scenarios exposes the model to a broader range of content requests which allows it to refine caching decisions based on diverse demand patterns. However, local high-demand conditions are not entirely dependent on MF-UAVs; each UAV continues learning independently to track sudden surges in requests. This ensures that the personalized Q-table remains highly reliable, as increased demand inherently leads to higher sampling, which is particularly beneficial for combinatorial bandits. This greater sampling rate allows UAVs to gain stronger confidence in learned content priorities, improve caching efficiency via high confidence *contribution factor* $\mathfrak{C}_{\mathbb{X},\mathbb{Y}}$ and reduce overall latency.

The results in Fig. 7 and 10 empirically validate these latency mitigation mechanisms. By leveraging controlled federated learning updates, demand-aware caching, and increased sampling during high-demand periods, the system effectively maintains low access delays under varying demand conditions that ensures robust performance in real-world UAV-assisted content dissemination scenarios.

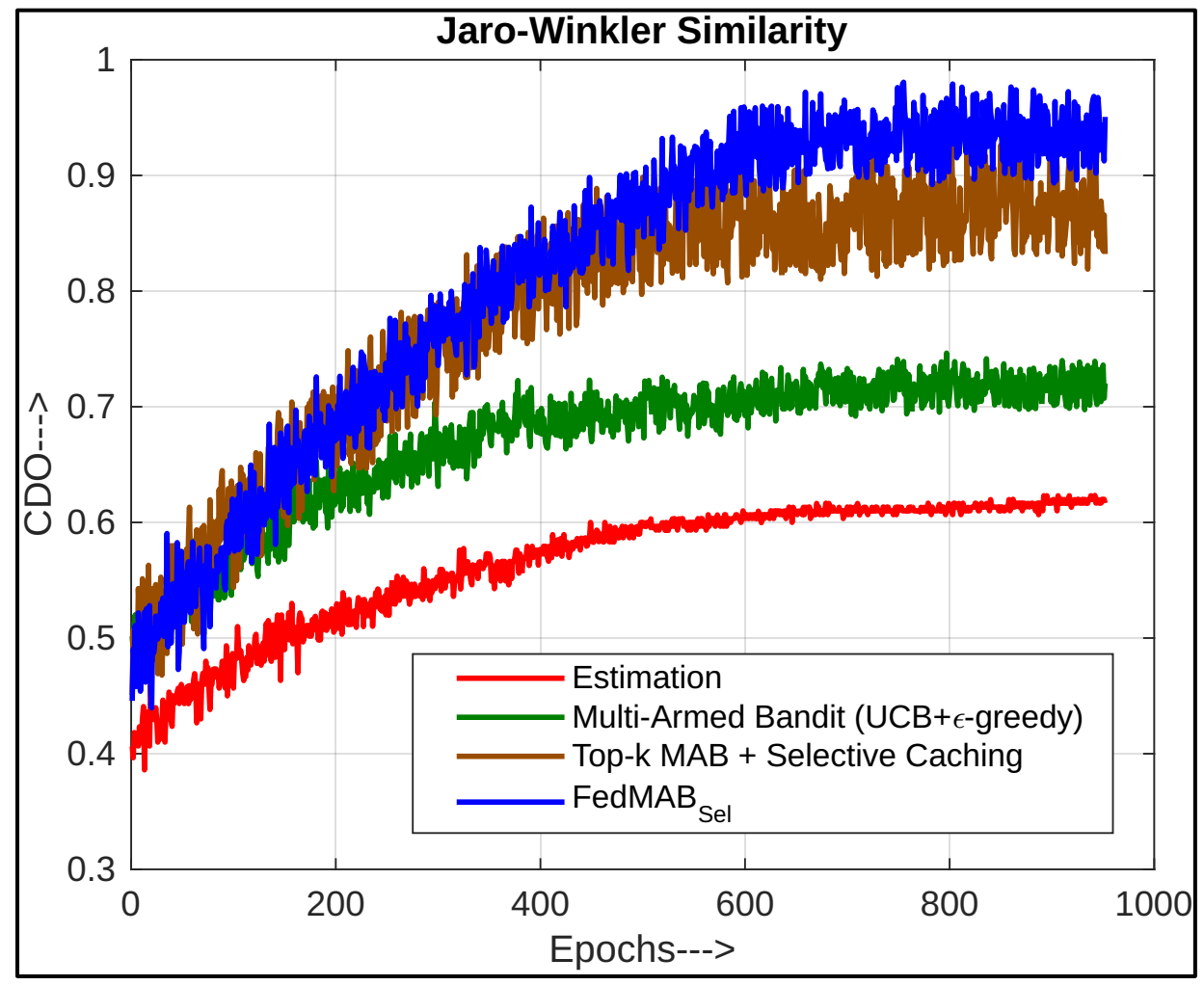

Fig. 11. Learnt cached content sequence's similarity with benchmark sequence

### 7.5 Cache Similarity of Learnt Sequence with Best Sequence

The effects of learning on the cached content sequence are demonstrated in Fig. 11. It plots Cache Distribution Optimality (*CDO*) of the cached content sequences for all the A-UAVs in terms of Jaro-Winkler Similarity (*JWS*). The key observation are as follows. First, the average $CDO$ between the benchmark caching sequence from cache pre-loading policy (see Section 5) and the cached content sequences learnt by the FedMAB agents at A-UAVs converge near 0.95, with relatively less variance with respect to its *Top-k* MAB predecessor. Physically, this represents higher degree of similarity after convergence, where 1 indicates complete similarity and 0 implies no similarity. Second, the cached contents improve over epochs as learning progresses. Lower $CDO$ values after the initial epochs signify that the A-UAVs have no *a priori* local or global content popularity information. As the MAB agents learn over multiple epochs of generated content requests, the cached contents in the A-UAVs become increasingly similar to the optimal caching sequence, which, in turn, improves the efficacy of FedMAB. Third, $CDO$ is an indirect representation of the storage segmentation factor ($\lambda$), which is used to decide the segment sizes according to cache pre-loading policies [26, 28]. A higher $CDO$ implies that, along with learning, the caching policy, the FedMAB agents learn to emulate the said segmentation behavior. Finally, the partial dissimilarity of the cached content sequence can be ascribed to the uncertainty (or regret) associated with the Q-values of contents with low popularity. Also, this leads to an oscillatory convergence of $CDO$ for the A-UAVs.

The impacts of selective caching at micro-ferrying UAVs can be distinctly seen in Fig 6. Selective caching at the MF-UAVs along with *Top-k* MAB caching agent at A-UAVs leads to a $CDO$ of nearly 0.9, although with a certain variance. Note that this depends on effective caching capacity of the MF-UAVs, which is dictated by the $TAD$s associated with content requests and the MF-UAVs visiting frequency at A-UAVs (refer Algorithm 3). The dependance of contents' Q-values on such information also adds to the post-convergence oscillation. Such oscillatory uncertainties are mitigated by the FedMAB, which enhances the value difference between in-demand and low demand contents, therefore improving the expect reward. To be noted that for the computation of $CDO$, the benchmark caching sequence is derived by considering the same effective caching capacity as the selective caching algorithm at the micro-ferrying UAVs.

## 8 CONCLUSION

In this paper, we propose a micro-UAV-assisted content dissemination system that learns caching policies on the fly without prior knowledge of content popularity. Two types of UAVs are introduced for content provisioning in disaster or war-stricken scenarios; anchor UAVs and micro-ferrying UAVs. Cache-enabled anchor UAVs are stationed at each stranded community of users to provide uninterrupted content delivery, while micro-ferrying UAVs act as content transfer agents between the anchor UAVs.

To overcome the limitations of existing caching methods, we introduce a decentralized Federated Multi-Armed Bandit (FedMAB) learning-based caching policy. This method leverages the collective intelligence of all A-UAVs to increase promptness in learning the caching policy, while reducing the redundant copies of the contents across the network. The policy at each A-UAV learns the caching decisions dynamically by maximizing an estimated multi-dimensional reward aimed at increasing both local and global content availability. Our results show that the FedMAB learning-based caching policy achieves approximately 88% of the maximum achievable content availability.

To further improve the Q-value estimates, we implement a Selective Caching Algorithm at the micro-ferrying UAVs. This method leverages shared information between anchor UAVs and micro-ferrying UAVs to further reduce the redundant content copies and provide a better estimate of the most popular content within a community. Combining selective caching at micro-ferrying UAVs with the FedMAB learning-based caching policy at anchor UAVs boosts content availability to approximately 94% of the maximum achievable level. With the proposed caching policies, a scaled-up micro-UAV-assisted network is shown to attain content availability close of the maximum achievable content availability.

*Discussion and Future Work*: Future work includes developing algorithms to handle time-varying content popularity and implementing adaptive trajectory planning to address operational unreliabilities of the UAVs. Additionally, it is necessary to explore methods for preserving the richness of information when converting multi-modal disaster data into smaller-sized formats to enhance effective content caching capacity.

While the proposed framework has been validated through simulations, real-world deployment would require addressing practical challenges that includes UAV coordination, wireless interference, and energy constraints. The decentralized structure of the system mitigates coordination complexity by allowing A-UAVs and MF-UAVs to operate autonomously which leverages federated learning to refine caching strategies without requiring continuous central control.

A transition to operational implementation would follow a phased approach. Initial deployment could involve small-scale testbed experiments, where UAVs execute FedMAB-based caching policies in controlled environments. This would allow for real-world validation of caching performance under dynamic network conditions. The next phase would involve field trials in real-world UAV-assisted networks, where interference, energy efficiency, and UAV mobility constraints could be evaluated. Future extensions could also explore hardware integration with UAV control algorithms which can ensure that caching decisions align with real-time flight dynamics.

By adopting a structured deployment roadmap, the proposed framework can be progressively refined for real-world applications while maintaining its efficiency in adaptive caching.